\documentclass[runningheads]{llncs}

\usepackage{eccv}

\usepackage{eccvabbrv}

\usepackage{graphicx}
\usepackage{booktabs}

\usepackage[accsupp]{axessibility}  

\usepackage[pagebackref,breaklinks,colorlinks,citecolor=eccvblue]{hyperref}

\usepackage{orcidlink}

\usepackage{subfloat}
\usepackage{multirow}
\usepackage{colortbl}
\begin{document}

\title{YYDS: Visible-Infrared Person Re-Identification with Coarse Descriptions} 

\titlerunning{Abbreviated paper title}

\author{
  Yunhao Du\inst{1} \and
  Zhicheng Zhao\inst{1,2,3} \and
  Fei Su\inst{1,2,3}
}
%
%

\institute{
  $^1$The school of Artificial Intelligence, Beijing University of Posts and Telecommunications \\
  $^2$Beijing Key Laboratory of Network System and Network Culture, Beijing, China\\
  $^3$Key Laboratory of Interactive Technology and Experience System Ministry of Culture and Tourism, Beijing, China\\
  {\tt\small \{dyh\_bupt,zhaozc,sufei\}@bupt.edu.cn} \\
}

\maketitle

\begin{abstract}
  Visible-infrared person re-identification (VI-ReID) is challenging due to considerable cross-modality discrepancies.
  Existing works mainly focus on learning modality-invariant features while suppressing modality-specific ones. %
  However, retrieving visible images only depends on infrared samples is an extreme problem because of the absence of color information.
  To this end, we present the \textbf{Refer-VI-ReID} settings, 
  which aims to match target visible images from both infrared images and coarse language descriptions 
  (e.g., ``a man with red top and black pants'') to complement the missing color information.
  To address this task, we design a Y-Y-shape decomposition structure, dubbed \textbf{YYDS}, to decompose and aggregate texture and color features of targets.
  Specifically, the text-IoU regularization strategy is firstly presented to facilitate the decomposition training, %
  and a joint relation module is then proposed to infer the aggregation. %
  Furthermore, the cross-modal version of k-reciprocal re-ranking algorithm is investigated, named \textbf{CMKR},
  in which three neighbor search strategies and one local query expansion method are explored to alleviate the modality bias problem of the near neighbors.
  We conduct experiments on SYSU-MM01, RegDB and LLCM datasets with our manually annotated descriptions.
  Both YYDS and CMKR achieve remarkable improvements over SOTA methods on all three datasets.
  Codes are available at \url{https://github.com/dyhBUPT/YYDS}.
\end{abstract}

\section{Introduction}
\label{sec:intro}

\begin{figure}[htbp]
    \centering
    \subfloat[VI-ReID]{\includegraphics[width=.25\textwidth]{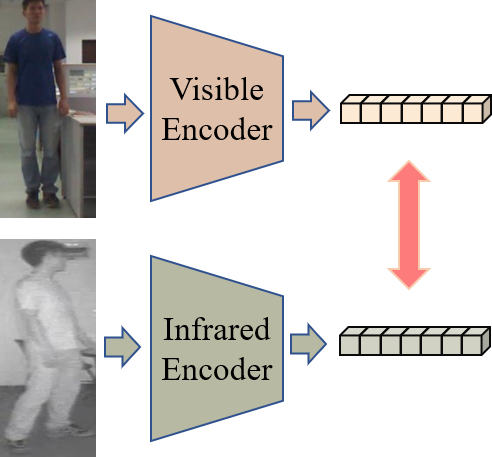}}
    \subfloat[TI-ReID]{\includegraphics[width=.35\textwidth]{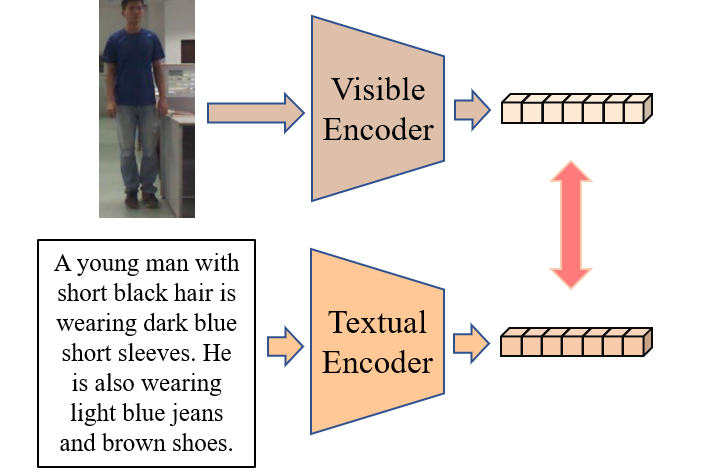}}
    \subfloat[Refer-VI-ReID]{\includegraphics[width=.25\textwidth]{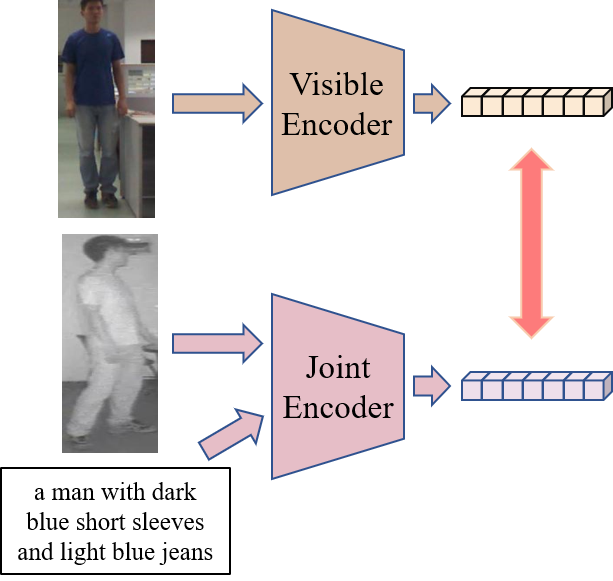}}
    \caption{
        \textbf{Comparison between different task settings.}
        (a) Visible-Infrared ReID.
        (b) Text-Image ReID.
        (c) Our proposed referring visible-infrared ReID.
    }
    \label{fig_1}
\end{figure}

Person re-identification (ReID) \cite{ye2021deep} aims to associate person images of the same identities across different cameras
and has significant research impacts on intelligent surveillance systems \cite{du2023strongsort, du2022pami}.
Benefiting from enormous amount of annotated data, deep learning based methods have achieved impressive success in recent years 
\cite{zheng2017person, li2014deepreid, zhong2018camera, sun2018beyond, luo2019strong, he2023fastreid, zhou2021learning, he2021transreid, li2023clip}.
However, their practical application is limited since only visible (RGB) modality is considered, which loses efficacy under low-light condition.

To meet the need for 24-hour surveillance systems, visible-infrared person re-identification (VI-ReID) is attracting increasing attentions.
Instead of performing matching within the visible modality, VI-ReID aims to retrieve visible targets with infrared probes \cite{wu2017rgb, lin2022learning, du2023video}.
Most existing VI-ReID methods perform \textit{shared feature learning}, which focus on embedding the features from different modalities into the same feature space
\cite{ye2020dynamic, ye2018hierarchical, ye2018visible, ye2020cross, liu2020parameter, chen2021neural, fu2021cm, ye2019bi, dai2018cross, wang2022optimal}.
Nevertheless, modality-specific information in both visible and infrared modalities are abandoned, including some discriminative details.
Differently, \textit{feature compensation learning} methods try to make up the missing modality-specific cues from one modality to another
\cite{wang2019rgb, wang2019learning, zhong2021grayscale, zhang2022modality, huang2021alleviating}.
However, it is an ill-posed problem to fill in the missing information in modality, e.g., color cues. 
This issue prompts us to consider a more sound strategy for information compensation.

Text-image person re-identification (TI-ReID) is one of the current research hotspots due to its wide application and flexibility
\cite{li2017person, zhang2018deep, wang2020vitaa, zheng2020dual, shao2022learning, jiang2023cross, yan2023clip, niu2023improving}.
It can search for the target pedestrian by the witness's language descriptions when visible probes are not available.
The success of TI-ReID proves that textual descriptions can provide discriminative information for person retrieval.
Inspired by this, we propose referring visible-infrared person re-identification (\textbf{Refer-VI-ReID}) settings,
which uses coarse descriptions, e.g., ``a man with dark blue short sleeves and light blue jeans'', to assist infrared images in retrieving visible images.
Fig.\ref{fig_1} illustrates the difference between VI-ReID, TI-ReID and Refer-VI-ReID.

\begin{figure}[t]
    \centering
    \includegraphics[width=.9\textwidth]{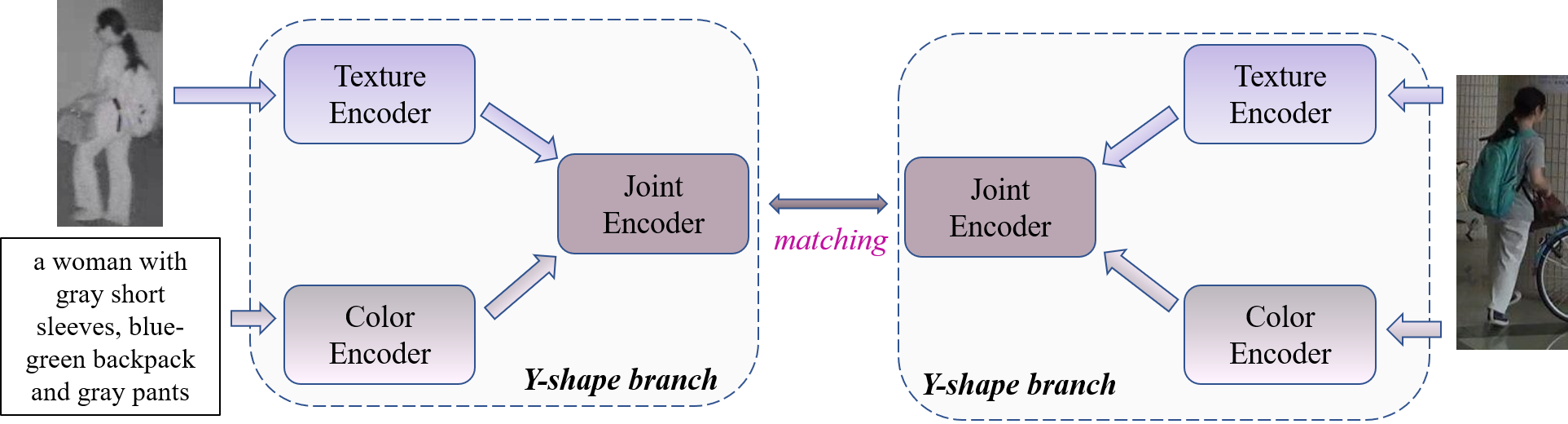}
    \caption{
        \textbf{Overview of our Y-Y-shape decomposition structure.}
    }
    \label{fig_2}
\end{figure}

The main challenge of Refer-VI-ReID lies in the misalignment between query modality (infrared and text) and gallery modality (visible).
The effectiveness of two-stream frameworks has been verified in previous cross-modal works 
\cite{chun2021probabilistic, wang2023multilateral, luo2022clip4clip, huang2023vop, du2022omg}.
In this paper, we extend this pipeline by designing a Y-Y-shape decomposition structure (\textbf{YYDS}) 
in the spirit of \textit{divide and conquer}, as shown in Fig.\ref{fig_2}.
Specifically, it includes two symmetric Y-shape branches to disentangle texture and color features from query and gallery samples respectively.
Previous disentanglement learning works tend to utilize auto-encoders \cite{kramer1991nonlinear} 
or GANs (Generative Adversarial Networks) \cite{goodfellow2014generative} to separate ID-discriminative factors and ID-excluded factors 
\cite{zheng2019joint, ge2018fd, zhang2019gait, li2020gait, choi2020hierarchical}, which suffer from difficult convergence and high training cost.
Some other works use extra clues (e.g., attributes, segmentation) to force the model to focus on identity information \cite{feng2023shape, chen2021explainable},
and don't take full advantage of the complementary features.
Differently, our Y-shape branch applies two separate encoders to extract complementary texture and color features individually.
Afterwards, they are dynamically aggregated by a joint encoder with asymmetric relation operation to predict the output features.
While training, texture and joint encoders are optimized with common ReID loss, 
and color encoders are optimized by Kullback-Leibler divergence loss with novel text-IoU regularization.
This design helps the network to extract both discriminative and complete representations without requiring adversarial or generative training.

$K$-reciprocal re-ranking \cite{zhong2017re} is widely used to revise the initial ranking list 
\cite{kalayeh2018human, wang2018resource, luo2019spectral, liu2020unity, tan2021incomplete, wang2022nformer, du2023video}.
It first searches for $k$-reciprocal nearest neighbors for each sample, and then embeds them into $k$-reciprocal features to calculate the Jaccard distance for re-ranking.
This strategy stands out from other re-ranking methods \cite{zhang2023graph, ouyang2021contextual, shen2021re} with its superior performance and unsupervised manner.
However, we find that it only brings limited performance improvements when applied to cross-modal tasks, e.g., VI-ReID.
Specifically, it is observed that neighbors are dominated by intra-modal samples (including many negative samples), while few cross-modal positive samples are included.
We refer to this problem as \textit{neighbor modality bias}, which is caused by the difference in intra-modal and inter-modal distance distributions.
Some previous works try to alleviate it by constraining the search domain of neighbors to the gallery set, 
which helps exclude intra-modal negative samples \cite{gao2021contextual, jia2020similarity, liang2021homogeneous}.
However, we find this \textit{constrained} strategy to be suboptimal because rich intra-modal context information is discarded.
To crack this nut, we propose two novel strategies, \textit{divided} and \textit{extended}, to introduce cross-modal positive neighbors while retaining intra-modal neighbors.
Furthermore, a modality-aware local query expansion (MA-LQE) strategy is designed to enhance the robustness of neighbor features.
Experiments demonstrate the superiority of our proposed method, dubbed (\textbf{CMKR}).

The main contributions can be summarized as follows:
\begin{itemize}
    \item[$\bullet$] We introduce the Refer-VI-ReID task setting, which assists visible-infrared retrieval with complementary textual information.
    \item[$\bullet$] We propose YYDS to perform texture and color feature disentanglement and aggregation to output discriminative and complete features.
    \item[$\bullet$] We design the CMKR algorithm, which extends k-reciprocal re-ranking to cross-modal scenarios and mitigates the effect of neighbor modality bias.
    \item[$\bullet$] Extensive experiments are conducted on SYSU-MM01, RegDB and LLCM to demonstrate the effectiveness of our methods.
\end{itemize}

\section{Related Work}
\label{sec:related}

\subsection{Visible-Infrared Person ReID}

Visible-infrared person ReID is challenging due to the cross-modality discrepancies between visible and infrared images.
Metric learning is widely used to narrow the gap between the two modalities by elaborately designing objective functions,
such as bi-directional dual-constrained top-ranking (BDTR) loss\cite{ye2018visible, ye2019bi}, hetero-center triplet loss \cite{liu2020parameter},
Margin MMD-ID loss \cite{jambigi2021mmd}, and so on \cite{ye2018hierarchical, hao2019hsme, hao2019dual, ling2020class, zhu2020hetero, hu2020cross, ye2020bi}.
Some other works focus on the network structure, e.g., partially shared two-stream framework \cite{ye2019modality, hao2021cross},
neural search based methods \cite{fu2021cm, chen2021neural}, adaptive style normalization \cite{wu2023style}, 
attention-based methods \cite{wei2020co, jiang2022cross, park2021learning}, etc.
Differently, generation-based methods utilize generative models \cite{kramer1991nonlinear, goodfellow2014generative, kingma2013auto}
to produce auxiliary samples \cite{wang2019rgb, wang2020cross, li2020infrared, fan2022modality, wei2021syncretic, zhang2021towards, qi2023generative, zhang2021rgb, kong2021dynamic, huang2021alleviating} 
or medial features \cite{zhang2022modality, yu2023modality}, or to realize disentanglement learning \cite{choi2020hi, pu2020dual, zhang2022fmcnet}.
Though their effectiveness, these methods suffer from high computational cost and difficult convergence while training.
Inspired by this, another series of works directly utilize data augmentation \cite{huang2022modality, kim2023partmix, wu2023style, ye2021channel, du2023enhanced}
or image transformation \cite{basaran2020efficient, ye2020visible, liu2021sfanet, zhang2023protohpe, li2023intermediary} strategies to generate diverse samples.

Recently, several works alleviate the cross-modal issues by introducing extra clues to learn discriminative representations.
Zhang \textit{et al.}\cite{zhang2020deep} manually annotated attribute labels as auxiliary information to train the model with attribute classification losses.
Zheng \textit{et al.}\cite{zheng2022progressive} further designed a progressive attribute embedding strategy to effectively fuse attribute information and visual information.
Chen \textit{et al.}\cite{chen2022structure} proposed to utilize key points to dynamically select discriminative appearance regions.
Feng \textit{et al.}\cite{feng2023shape} decorrelated modality-shared features into shape-related and shape-erased features under the guidance of human parsing maps.
Different from previous works, we present a novel Y-Y-shape structure to assist cross-modal retrieval with coarse descriptions, which yields a wider range of applications.

\subsection{Re-ranking}

Re-ranking is commonly used in various retrieval tasks, which aims to revise the initial ranking list with sample-to-sample similarity.
Jegou \textit{et al.}\cite{jegou2007contextual, jegou2008accurate} presented an iterative approach to reduce the neighborhood non-reversibility rate.
Qin \textit{et al.}\cite{qin2011hello} formally defined the concept of $k$-reciprocal nearest neighbors, 
and utilized two rounds of query expansion to circumvent the non-reciprocity of near neighbor relationships.
Leng \textit{et al.}\cite{leng2013bidirectional, leng2015person} proposed the bi-directional ranking mechanism and considered both content and context similarity.
Similarity, Garc\'ia \textit{et al.}\cite{garcia2015person} applied content and context information to remove the visual ambiguity.
Ye \textit{et al.}\cite{ye2015coupled, ye2016person} pursued accurate retrieval results by aggregating multiple ranking lists.
Bai \textit{et al.}\cite{bai2016sparse} designed an elaborate pipeline called the sparse contextual activation (SCA),
which measures the dissimilarity between samples by the Jaccard distance of neighbor features.
Zhong \textit{et al.}\cite{zhong2017re} extended from SCA by integrating the idea of $k$-reciprocal nearest neighbors.
Yu \textit{et al.}\cite{yu2017divide} exploited the diversity of high-dimensional features in the spirit of ``divide and fuse''.
Sarfraz \textit{et al.}\cite{sarfraz2018pose} directly calculated the averaged distance of expanded neighbors as the final measurement.
Guo \textit{et al.}\cite{guo2018density} introduced a soft version of $k$-reciprocal nearest neighbors with a density-adaptive kernel technique.
Peng \textit{et al.}\cite{peng2019re} assumed that the probe image and its neighbors lie in a locally linear manifold and obtained neighbor features with linear reconstruction.
Chen \textit{et al.}\cite{chen2023jaccard} improved Jaccard distance in an camera-aware manner.

These re-ranking methods have shown significantly improved performance in single-modal retrieval tasks.
Nevertheless, limited performance is observed when directly applying them to cross-modal scenarios caused by the neighbor modality bias problem.
Existing methods \cite{gao2021contextual, jia2020similarity, liang2021homogeneous} tackled this problem by searching neighbors only in the gallery set.
We argue that this ``constrained'' strategy is suboptimal because it ignores substantial context information in the query set.
In this paper, we propose a novel cross-modal k-reciprocal re-ranking algorithm to avoid such a dilemma.

\section{Method}
\label{sec:method}

\subsection{Preliminary}

Let us take $\mathcal{X}^v = \{x_n^v\}_{n=1}^{N^v}$ and $\mathcal{X}^i = \{x_n^i\}_{n=1}^{N^i}$ to denote visible and infrared images,
where $x_n^v$ and $x_n^i$ are images, and $N^v$ and $N^i$ are the number of visible and infrared images.
The corresponding identity labels are defined as $\mathcal{Y}^v = \{y_n^v\}_{i=1}^{N^v}$ and $\mathcal{Y}^i = \{y_n^i\}_{n=1}^{N^i}$.
Coarse descriptions $\mathcal{X}^t = \{x_n^t\}_{n=1}^{N^i}$ are manually annotated for each infrared image, where samples of the same identity share the same description.
The goal of Refer-VI-ReID is to match samples across modalities 
by learning modality invariant representations $\mathcal{F}^v = \{f_n^v\}_{n=1}^{N^v}$ and $\mathcal{F}^i = \{f_n^i\}_{n=1}^{N^i}$.
While inference, given a pair of query infrared image and coarse description $q = \{q^i, q^t\} \in \mathcal{Q}$, 
the retrieval ranking list $\mathcal{RL}(q) = \{g_1^v, g_2^v,...,g_{N_\mathcal{G}}^v | g_n^v \in \mathcal{G}\}$ is generated 
according their feature similarities with the visible gallery set $\mathcal{G}$.

\begin{figure}[t]
    \centering
    \includegraphics[width=.9\textwidth]{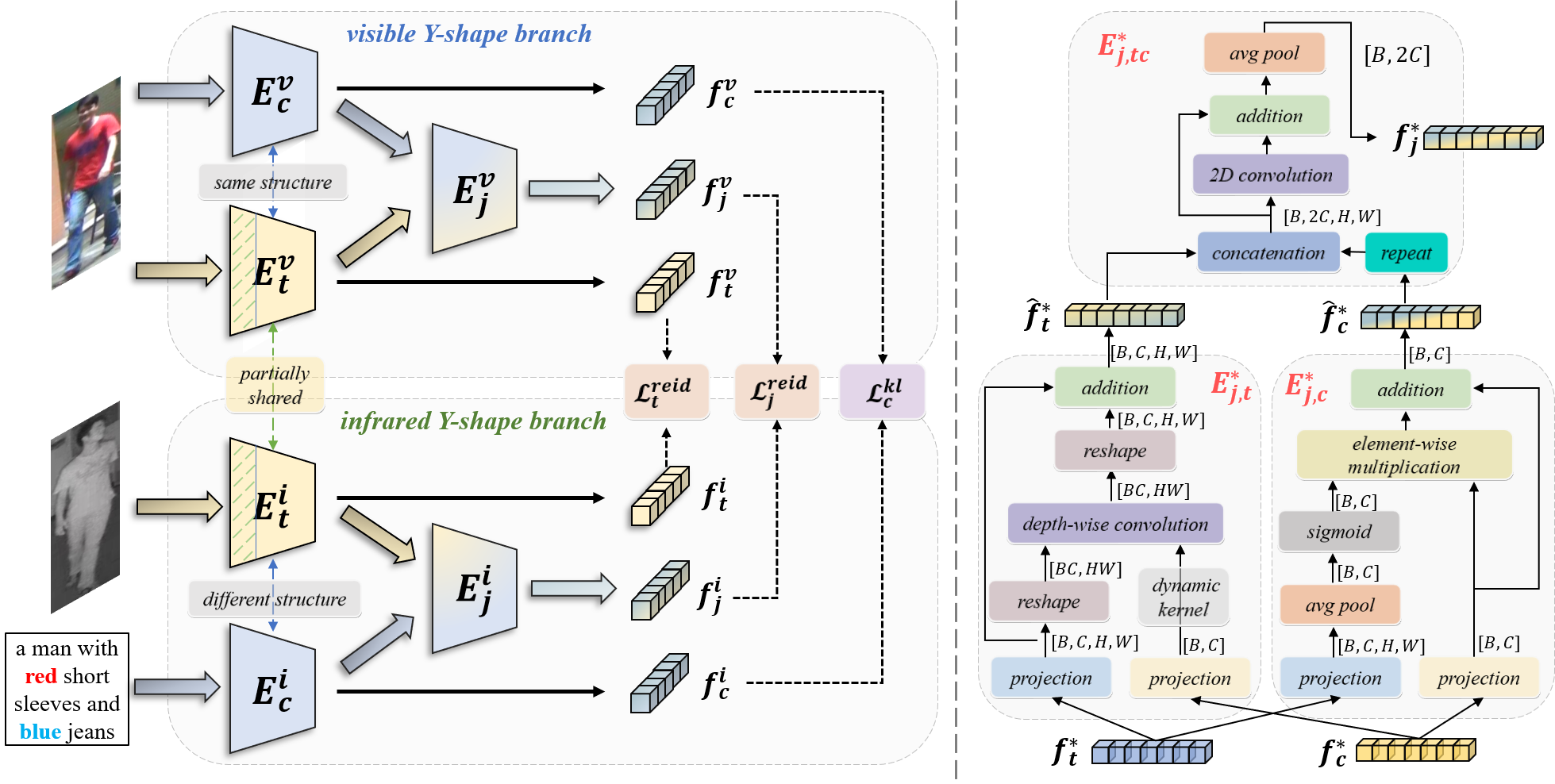}
    \caption{
        \textbf{Left:} The framework of YYDS, which includes a visible Y-shape branch and an infrared Y-shape branch.
        Each branch consists of a color encoder $E_c^*$, a texture encoder $E_t^*$ and a joint relation module (JRM) $E_j^*$.
        The two $E_t^*$ partially share weights to eliminate modality-specific information.
        During training, the overall framework is optimized by two ReID loss $L_t^{reid}$, $L_j^{reid}$ and one KL divergence loss $L_c^{kl}$ with text-IoU regularization.
        \textbf{Right:} The details of JRM, which is composed of texture-centered relation block $E_{j,t}^*$, 
        color-centered relation block $E_{j,c}^*$ and joint relation block $E_{j,tc}^*$.
        $B$ is the batch size, $C$ is the channel dimension, and $H,W$ is the size of feature map.
    }
    \label{fig_3}
\end{figure}

\subsection{YYDS}

\subsubsection{Overview}
\
\newline
\newline
\indent 
It is assumed that the identity discriminative features in visible person images include color features and texture features.
However, infrared images only contain texture information and lack sufficient color information.
Inspired by this, we propose to utilize language descriptions to complete the missing color information with a novel color-texture disentanglement framework.
Fig.\ref{fig_3} (left) illustrates the designed Y-Y-shape decomposition structure (YYDS), which primarily comprises two branches,
i.e., visible Y-shape branch and infrared Y-shape branch.
The two branches share the similar structure, including a color encoder $E_c^*$, a texture encoder $E_t^*$ and a joint relation module (JRM) $E_j^*$,
where $* \in {v, i}$ represents image modalities.

Specifically, it takes a mini-batch of triplets $\{x_n^v, x_n^i, x_n^t\}_{n=1}^{N_B}$ as inputs, where $N_B$ is the batch size.
For simplification, the subscript $n$ is omitted in the following descriptions.
The two texture encoders $E_t^v$ and $E_t^i$ are utilized to extract texture embeddings $f_t^* = E_t^*(x^*), * \in \{v,i\}$,
which utilizes ResNet-50 \cite{he2016deep} as the backbone and the last stride is set to 1.
Following previous VI-ReID works \cite{ye2019modality, hao2021cross}, 
the first convolutional blocks in $E_t^v$ and $E_t^i$ don't share weights to capture modality-specific low-level features,
and the parameters of deeper blocks are shared to learn high-level texture features.
As for color encoders, the $E_c^v$ share the same structure with $E_t^v$ with unshared parameters to capture color features from visible images $f_c^v = E_c^v(x^v)$.
Differently, $E_c^i$ takes descriptions $x^t$ as input $f_c^i = E_c^i(x^t)$, which applies RoBERTa \cite{liu2019roberta} as the backbone.
Afterwards, joint relation modules $E_j^*, * \in \{v, i\}$ are designed to dynamically aggregate the decomposed color embeddings $f_c^*$ and $f_t^*$,
and output the joint embeddings $f_j^* = E_j^*(f_c^*, f_t^*)$ (details in Sec \ref{sec:JRM}).

During training, the three pairs of encoders $\{E_c^v, E_c^i\}$, $\{E_t^v, E_t^i\}$ and $\{E_j^v, E_j^i\}$ are optimized 
by one Kullback-Leibler (KL) divergence loss $\mathcal{L}_c^{kl}$ and two common ReID losses $\mathcal{L}_t^{reid}$ and $\mathcal{L}_j^{reid}$ respectively (details in Sec \ref{sec:loss}).
While inference, the triplet of embeddings $\{f_c^*, f_t^*, f_j^*\}$ are concatenated along the channel dimension to obtain the final features $f^*, * \in \{v, i\}$.

\subsubsection{Joint Relation Module}
\label{sec:JRM}
\
\newline
\newline
\indent
The details of joint relation modules (JRM) $E_j^*$ are shown in Fig.\ref{fig_3} (right), which consists of three relation blocks $E_{j,t}^*$, $E_{j,c}^*$ and $E_{j,tc}^*$.
The motivation is to first perform mutual modulation between color and texture embeddings $f_c^*$ and $f_t^*$ 
with color-centered relation block $E_{j,c}^*$ and texture-centered relation block $E_{j,t}^*$,
and then the two modulated features are aggregated with the joint relation block $E_{j, tc}^*$.

Because of the great discrepancy between $f_c^*$ and $f_t^*$, the mutual modulation procedure is designed in an asymmetric manner,
i.e., \textit{channel-level modulation} and \textit{spatial-level modulation}.
For $E_{j,c}^*$, $f_c^*$ and $f_t^*$ are first embedded by two projection heads.
Then, the channel modulation weights $w_t^*$ are generated by projected texture features $\overline{f}_t^*$ with an average pooling layer and a sigmoid layer:
\begin{equation}
    w_t^* = sigmoid(pool(\overline{f}_t^*)).
\end{equation}
Then the projected color features $\overline{f}_c^*$ is modulated by multiplying $w_t^*$ with a residual addition:
\begin{equation}
    \hat{f}^*_c = w_t^* \times \overline{f}_c^* + \overline{f}_c^*,
\end{equation}
where $\times$ denotes element-wise multiplication. 
As to $E_{j, t}^*$, $f_c^*$ and $f_t^*$ are also passed through two projection heads firstly to output $\tilde{f}_c^*$ and $\tilde{f}_t^*$.
Different from $f_c^*$, the texture features $f_t^*$ contain rich spatial contextual clues.
Inspired by this, the dynamic convolution is performed to modulate local spatial features in $\tilde{f}_t^*$ under the guidance of $\tilde{f}_c^*$.
Specifically, we take $\tilde{f}_c^*$ as dynamic kernels to convolute with $\tilde{f}_t^*$ by depth-wise convolution \cite{sifre2014rigid, howard2017mobilenets, chollet2017xception}.
Then a residual addition is followed for optimization stability:
\begin{equation}
    \hat{f}_t^* = DWConv(\tilde{f}_t^*; \tilde{f}_c^*) + \tilde{f}_t^*,
\end{equation}
where $DWConv(\cdot; \cdot)$ represents depth-wise convolution.

After obtaining modulated features $\hat{f}^*_c$ and $\hat{f}_t^*$, the joint relation block $E_{j, tc}^*$ is introduced to conduct aggregation.
$\hat{f}^*_c$ is first repeated in spatial dimensions and then is concatenated with $\hat{f}_t^*$ along the channel dimension to output $\hat{f}^*_{tc}$.
Afterwards, a 2D convolution layer with residual addition and an average pooling layer are utilized to extract the joint embeddings $f_j^*$:
\begin{equation}
    f_j^* = pool(conv(\hat{f}^*_{tc}) + \hat{f}^*_{tc}).
\end{equation}

In conclusion, the two disentangled embeddings $f_c^*$ and $f_t^*$ are first modulated by each other in channel and spatial dimensions respectively,
and then are aggregated to output the final discriminative and complete embeddings $f_j^*$.

\begin{figure}[t]
    \centering
    \includegraphics[width=.7\textwidth]{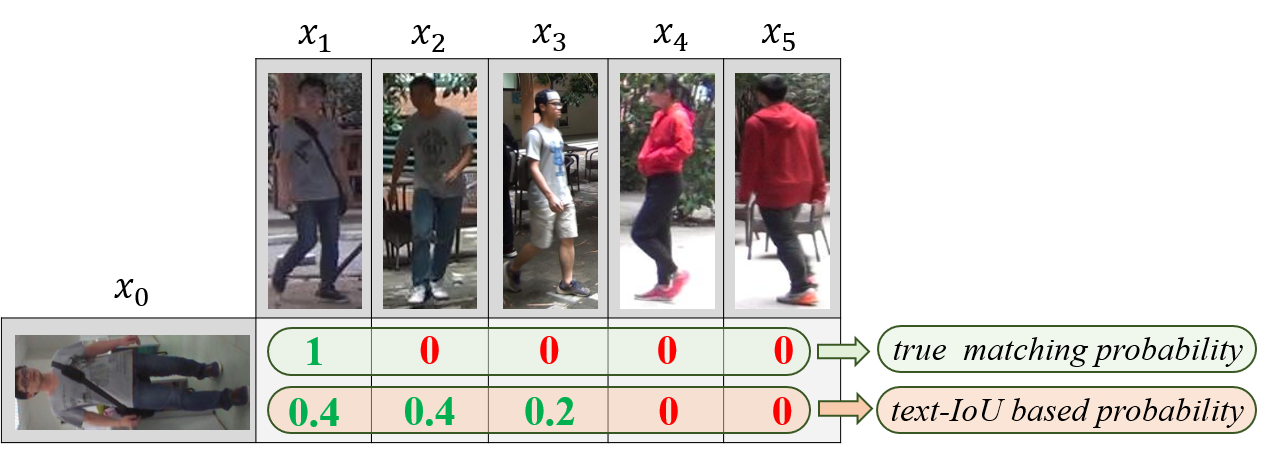}
    \caption{
        \textbf{The illustration of text-IoU regularization.}
        Only $x_1$ shares the same identity with $x_0$, but $x_2$ and $x_3$ share similar color clues with $x_0$.
    }
    \label{fig_iou}
\end{figure}

\subsubsection{Training Loss}
\label{sec:loss}
\
\newline
\newline
\indent
Commonly used ReID losses (cross-entropy loss and triplet loss) 
$\mathcal{L}^{reid}_t$ and $\mathcal{L}^{reid}_j$ are utilized to train texture encoders $E_t^*$ and JRM $E_j^*$ respectively.
Please note that our framework can adapt to various VI-ReID backbones and losses.
In this paper, we take DEEN \cite{zhang2023diverse} as baseline, and their proposed CPM loss and orthogonal loss are also included in our ReID losses.

As to color encoders, the distribution matching loss based on Kullback-Leibler (KL) divergence is applied 
as in previous text-image ReID works \cite{zhang2018deep, jiang2023cross}:
\begin{equation}
    \mathcal{L}^{kl}_c = {1 \over N_B} \sum_{n=1}^{N_B} \sum_{m=1}^{N_B} p_{n,m} log({p_{n,m} \over {q_{n, m} + \epsilon}}), \label{kl}
\end{equation}
where $p_{n,m}$ is the probability of matching pairs $\{f_n, f_m\}$, $q_{n, m}$ is the true matching probability, and $\epsilon=1e^{-8}$ is applied to avoid numerical problems.
In implementation, four pairs of features are used to calculate the KL loss, i.e., $\{f^v_c, f^i_c\}$, $\{f^i_c, f^v_c\}$, $\{f^v_c, f^v_c\}$ and $\{f^i_c, f^i_c\}$.

However, different identities may share similar color information.
Fig.\ref{fig_iou} illustrates an example, where $x_0$ has the same identity only with $x_1$, but are also dressed in similar colors with $x_2$ and $x_3$.
The color encoders would be confused while being optimized by true matching probability $q_{0,*} = [1,0,0,0,0]$ as in Eq.\ref{kl}.
To circumvent such issue, we propose a text-IoU regularization strategy, which replace the original $q_{n,m}$ with text-IoU based probability $\hat q_{n,m}$.
Concretely, given a description $x^t_n$, NLTK \cite{bird2009natural} is utilized to extract color words $c_n = \{w_1, ..., w_{N_n}\}$.
Then, the text-IoU between samples $n$ and $m$ is defined by the intersection over union of their color word sets:
\begin{equation}
    IoU_{n,m} = {{|c_n \cap c_m|} \over {|c_n \cup c_m|}},
\end{equation}
where $|\cdot|$ is the cardinality of sets.
Finally, the normalized text-IoU is utilized as the regularization probability $\hat q_{n,m}$ to replace $q_{n,m}$ in Eq.\ref{kl},
which equals to $[0.4, 0.4, 0.2, 0, 0]$ in the example in Fig.\ref{kl}.

Overall, the objective function of our YYDS can be summarized as follows:
\begin{equation}
    \mathcal{L}_{total} = \lambda_t \mathcal{L}^{reid}_t + \lambda_j \mathcal{L}^{reid}_j + \lambda_c \mathcal{L}^{kl}_c. \label{loss_all}
\end{equation}

\begin{figure}[htbp]
    \centering
    \includegraphics[width=.8\textwidth]{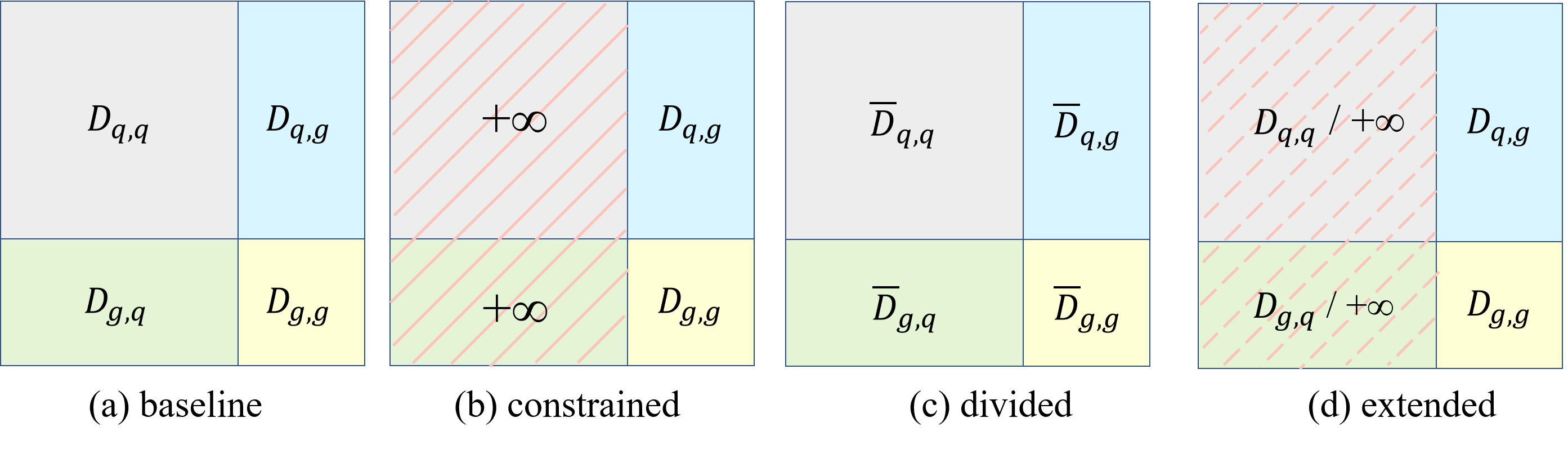}
    \caption{
        \textbf{The illustration of different neighbor strategies from the perspective of distance matrices.}
        (a) The baseline method searches for neighbors based on the original distance matrix.
        (b) The constrained strategy limits the search domain to the gallery set.
        (c) The divided strategy separately normalizes the four submatrices.
        (d) The extended strategy integrates the baseline and constrained strategies.
    }
    \label{fig_dist}
\end{figure}

\subsection{CMKR}

\subsubsection{Overview}
\
\newline
\newline
\indent
The $k$-reciprocal re-ranking algorithm \cite{zhong2017re} is widely utilized to refine the initial ranking list in the testing stage.
Given $N_q$ query images and $N_g$ gallery images, the original distance matrix $D_{ori} \in \mathbb R^{(N_q + N_g) \times (N_q + N_g)}$ can be represented as the concatenation of four submatrices:
\begin{equation}
    D_{ori} = \begin{bmatrix} D_{q,q} & D_{q,g} \\ D_{g,q} & D_{g,g} \end{bmatrix}, \label{eq_distance}
\end{equation}
where $D_{q,q} \in \mathbb R^{N_q \times N_q}$ is the distance between all query images, and the others are similar.
The $k$-reciprocal nearest neighbors of query $q_i$ are defined as:
\begin{equation}
    R(q_i, k_1) = \{g_j | (g_j \in N(q_i, k_1)) \land (q_i \in N(g_j, k_1)) \}, \label{eq_neighbor}
\end{equation}
where $N(q_i, k_1)$ is the $k$-nearest neighbors of $q_i$ searched based on $D_{ori}$.
For ease of calculation, $R(q_i, k_1)$ is transformed to neighbor features $V_i \in \mathbb R^{(N_q + N_g)}$, which is further refined by the local query expansion (LQE) strategy \cite{chum2007total, gordo2017end, radenovic2018fine}:
\begin{equation}
    \tilde {V}_i = {1 \over k_2} \sum_{q_j \in N(q_i, k_2)} V_j. \label{eq_neighbor_feat}
\end{equation}
Finally, the Jaccard distance based on refined neighbor features are calculated to perform re-ranking:
\begin{equation}
    d_{Jacc}(q_i, g_j) = 1 - {\sum_{n=1}^{N_q+N_g} min(\tilde{V}_{i,n}, \tilde{V}_{j,n}) \over \sum_{n=1}^{N_q+N_g} max(\tilde{V}_{i,n}, \tilde{V}_{j,n})},
\end{equation}
where $\tilde{V}_{i,n}$ is the $n$-th element of $\tilde{V}_i$.

\subsubsection{Neighbor Strategy}
\
\newline
\newline
\indent
The performance of re-ranking is greatly affected by the accuracy of searched neighbors.
However, in cross-modal scenarios, neighbors tend to be dominated by intra-modal positive and negative samples, including very few cross-modal samples.
To solve this neighbor modality bias problem, we study three neighbor strategies, i.e., \textit{constrained}, \textit{divided} and \textit{extended} strategies.

\noindent \textbf{Constrained.}
An intuitive solution is to constrain the neighbor search domain, as done in \cite{gao2021contextual, jia2020similarity, liang2021homogeneous}.
Specifically, define $N_{q \rightarrow g}(q_i, k_1)$ as the cross-modal neighbors of $q_i$ among gallery sets, and the constrained neighbors $R_{q \rightarrow g}$ is:
\begin{equation}
    R_{q \rightarrow g}(q_i, k_1) = \{ g_j | (g_j \in N_{q \rightarrow g}(q_i, k_1)) \land (q_i \in N_{g \rightarrow q}(g_j, k_1))  \}, \label{eq_cross_neighbor}
\end{equation}
which is utilized to replace the original neighbors $R(q_i, k_1)$ in Eq.\ref{eq_neighbor}.
However, we argue that this strategy is suboptimal because it retains cross-modal samples at the expense of losing intra-modal contextual information,
which is solved by the following two strategies.

\noindent \textbf{Divided.}
The neighbor modality bias problem is mainly caused by the distribution inconsistency between intra-modal distances $D_{q,q} / D_{g,g}$ and cross-modal distances $D_{q, g} / D_{g, q}$.
Inspired by this, the divided strategy conducts row min-max normalization to these four submatrices separately, and then concatenates them into a new distance matrix:
\begin{equation}
    D_{div} = \begin{bmatrix} \overline{D}_{q,q} & \overline{D}_{q,g} \\ \overline{D}_{g,q} & \overline{D}_{g,g} \end{bmatrix},
\end{equation}
where $\overline{D}_{*, *}$ is the row-normalized $D_{*, *}$.
Then, $D_{div}$ is applied to replace $D_{ori}$ in Eq.\ref{eq_distance} for neighbor search. 

\noindent \textbf{Extended.}
Different from the $divided$ strategy which operates on the distance matrix, the extended directly extends the original neighbor set by introducing cross-modal neighbors.
Specifically, it takes the union of $R(q_i, k_1)$ in Eq.\ref{eq_neighbor} and $R_{q \rightarrow g}(q_i, k_1)$ in Eq.\ref{eq_cross_neighbor} as the final neighbor set:
\begin{equation}
    R_{ext}(q_i, k_1) = R(q_i, k_1) \cup R_{q \rightarrow g}(q_i, k_1),
\end{equation}
which is used to calculate the neighbor features.

Fig.\ref{fig_dist} illustrates the comparison between baseline and the three proposed neighbor strategies from the perspective of distance matrices.
Experiments will demonstrate the superiority of the proposed neighbor strategies over the baseline method.

\subsubsection{MA-LQE}
\
\newline
\newline
\indent
The refined neighbor features $\tilde V_i$ in Eq.\ref{eq_neighbor_feat} also suffer from the neighbor modality bias problem, because $N(q_i, k_2)$ is mainly composed of intra-modal samples.
To alleviate this issue, we propose modality-aware local query expansion (MA-LQE) method, which searches for intra-modal and inter-modal neighbors separately to refine the neighbor features:
\begin{equation}
    \hat{V}_i = {1 \over k_2} \sum_{q_j \in \hat{N}(q_i, k_2, k_3)} V_j, \label{eq_neighbor_feat_2}
\end{equation}
where $\hat{N}(q_i, k_2, k_3) = N_{q \rightarrow q}(q_i, k_2-k_3) \cap N_{q \rightarrow g}(q_i, k_3)$ explicitly integrates $(k_2 - k_3)$ intra-modal neighbors and $k_3$ cross-modal neighbors.

\begin{table*}[p]
    \renewcommand{\arraystretch}{1.2}
    \begin{center}
        \caption{
            \textbf{Comparison with the state-of-the-art VI-ReID methods on SUSY-MM01 under the single-shot protocol.}
            ``Epoch'' represents the training epochs.
        }
        \label{table_sota_1}
        \resizebox{.95\textwidth}{!}{
        \begin{tabular}{cl|c|c|c|c|c|c|c|c|c|c}
            \toprule[1pt]
            & \multirow{2}*{\textbf{Method}} & \multirow{2}*{\textbf{Reference}} & \multirow{2}*{\textbf{Epoch}}
            & \multicolumn{4}{c|}{\textbf{All Search}} & \multicolumn{4}{c}{\textbf{Indoor Search}} \\
            \cline{5-12}
            & ~ & ~ & ~ & \textbf{R-1} & \textbf{R-10} & \textbf{R-20} & \textbf{mAP} & \textbf{R-1} & \textbf{R-10} & \textbf{R-20} & \textbf{mAP} \\
            \hline
            & cmGAN\cite{dai2018cross}          & IJCAI2018 & 2000 & 26.97 & 67.51 & 80.56 & 27.80 & 31.63 & 77.23 & 89.18  & 42.19 \\
            & AlignGAN\cite{wang2019rgb}        & ICCV2019  & 250  & 42.40 & 85.00 & 93.70 & 40.70 & 45.90 & 87.60 & 94.40  & 54.30 \\
            & MHM\cite{yang2020mining}          & AAAI2020  & 30   & 35.90 & 73.00 & 86.10 & 38.00 &   -   &   -   &   -    &   -   \\
            & JSIA\cite{wang2020cross}          & AAAI2020  & 650  & 38.10 & 80.70 & 89.90 & 36.90 & 43.80 & 86.20 & 94.20  & 52.90 \\
            & XIV\cite{li2020infrared}          & AAAI2020  & 120  & 49.92 & 89.79 & 95.96 & 50.73 &   -   &   -   &   -    &   -   \\
            & DDAG\cite{ye2020dynamic}          & ECCV2020  & 80   & 54.75 & 90.39 & 95.81 & 53.02 & 61.02 & 94.08 & 98.41  & 67.98 \\
            & LbA\cite{park2021learning}        & ICCV2021  & 80   & 55.41 &   -   &   -   & 54.14 & 58.46 &   -   &   -    & 66.33 \\
            & CM-NAS\cite{fu2021cm}             & ICCV2021  & 120  & 61.99 & 92.87 & 97.25 & 60.02 & 67.01 & 97.02 & 99.32  & 72.95 \\
            & CAJ\cite{ye2021channel}           & ICCV2021  & 100  & 69.88 & 95.71 & 98.46 & 66.89 & 76.26 & 97.88 & 99.49  & 80.37 \\
            & MPANet\cite{wu2021discover}       & CVPR2021  & 140  & 70.58 & 96.21 & 98.80 & 68.24 & 76.74 & 98.21 & 99.57  & 80.95 \\
            & SFANet\cite{liu2021sfanet}        & TNNLS2021 & 80   & 65.74 & 92.98 & 97.05 & 60.83 & 71.60 & 96.60 & 99.45  & 80.05 \\
            & MSCLNet\cite{zhang2022modality}   & ECCV2022  & 200  & 76.99 & 97.63 & 99.18 & 71.64 & 78.49 & 99.32 & 99.91  & 81.17 \\
            & PAENet\cite{zheng2022progressive} & MM2022    & 80   & 74.22 & 99.03 & 99.97 & 73.90 & 78.04 & 99.58 & 100.00 & 83.54 \\
            & DEEN\cite{zhang2023diverse}       & CVPR2023  & 150  & 74.70 & 97.60 & 99.20 & 71.80 & 80.30 & 99.00 & 99.80  & 83.30 \\
            & CAL\cite{wu2023learning}          & ICCV2023  & 100  & 74.66 & 96.47 &   -   & 71.73 & 79.69 & 98.93 &   -    & 83.68 \\
            & SAAI\cite{fang2023visible}        & ICCV2023  & 160  & 75.90 &   -   &   -   & 77.03 & 83.20 &   -   &   -    & 88.01 \\
            & MUM\cite{yu2023modality}          & ICCV2023  & 90   & 76.24 & 97.84 &   -   & 73.81 & 79.42 & 98.09 &   -    & 82.06 \\
            & DARD\cite{wei2023dual}            & TIFS2023  & 300  & 69.33 & 94.32 & 97.52 & 65.65 & 77.21 & 98.32 & 99.18  & 81.91 \\
            & TransVI\cite{chai2023dual}        & TCSVT2023 & 80   & 71.36 & 96.77 & 98.26 & 68.63 & 77.40 & 98.69 & 99.82  & 81.31 \\
            & TMD\cite{lu2023tri}               & TMM2023   & 80   & 73.92 & 96.29 & 98.76 & 67.76 & 81.16 & 98.87 & 99.68  & 78.88 \\
            & STAR\cite{wu2023style}            & TMM2023   & 100  & 76.07 & 97.76 &   -   & 72.73 & 83.47 & 99.04 &   -    & 85.76 \\ 
            \hline
            & Baseline                   &      ours     &     80      & 72.36 & 96.56 & 99.03 & 68.24 & 78.50 & 98.45 & 99.50 & 82.06 \\
            & \textbf{YYDS}              & \textbf{ours} & \textbf{80} & \textbf{85.54} & \textbf{99.30} & \textbf{99.78} & \textbf{81.64} 
                                                                       & \textbf{89.13} & \textbf{99.66} & \textbf{99.96} & \textbf{91.00} \\
            & \textbf{YYDS+CMKR}           & \textbf{ours} & \textbf{80} & \textbf{95.51} & \textbf{99.68} & \textbf{99.80} & \textbf{93.77} 
                                                                       & \textbf{98.59} & \textbf{99.64} & \textbf{100.00} & \textbf{98.41} \\
            \bottomrule[1pt]
        \end{tabular}
        }
    \end{center}
    \begin{center}
        \caption{
            \textbf{Comparison with the state-of-the-art VI-ReID methods on harsh lighting datasets RegDB and LLCM under the infrared-to-visible protocol.}
        }
        \label{table_sota_2}
        \resizebox{.90\textwidth}{!}{
        \begin{tabular}{cl|c|c|c|c|c|c|c|c|c}
            \toprule[1pt]
            & \multirow{2}*{\textbf{Method}} & \multirow{2}*{\textbf{Reference}}
            & \multicolumn{4}{c|}{\textbf{RegDB}} & \multicolumn{4}{c}{\textbf{LLCM}} \\
            \cline{4-11}
            & ~ & ~ & \textbf{R-1} & \textbf{R-10} & \textbf{R-20} & \textbf{mAP} & \textbf{R-1} & \textbf{R-10} & \textbf{R-20} & \textbf{mAP} \\
            \hline
            & DDAG\cite{ye2020dynamic}    & ECCV2020  & 68.1 & 85.2 & 90.3 & 61.8 & 41.0 & 73.4 & 81.9 & 49.6 \\
            & LbA\cite{park2021learning}  & ICCV2021  & 67.5 &  -   &  -   & 72.4 & 44.6 & 78.2 & 86.8 & 53.8 \\
            & AGW\cite{ye2021deep}        & TPAMI2021 &  -   &  -   &  -   &  -   & 46.4 & 77.8 & 85.2 & 54.8 \\
            & CAJ\cite{ye2021channel}     & ICCV2021  & 84.8 & 95.3 & 97.5 & 77.8 & 48.8 & 79.5 & 85.3 & 56.6 \\
            & DART\cite{yang2022learning} & CVPR2022  & 82.0 &   -  &   -  & 73.8 & 52.2 & 80.7 & 87.0 & 59.8 \\
            & MMN\cite{zhang2021towards}  & MM2021    & 87.5 & 96.0 & 98.1 & 80.5 & 52.5 & 81.6 & 88.4 & 58.9 \\
            & DEEN\cite{zhang2023diverse} & CVPR2023  & 89.5 & 96.8 & 98.4 & 83.4 & 54.9 & 84.9 & 90.9 & 62.9 \\
            \hline
            & Baseline      &     ours      & 89.1 & 96.8 & 98.5 & 81.8 & 56.5 & 85.3 & 91.3 & 63.2 \\
            & \textbf{YYDS} & \textbf{ours} & \textbf{90.2} & \textbf{97.3} & \textbf{98.8} & \textbf{83.5} 
                                            & \textbf{58.2} & \textbf{87.2} & \textbf{92.6} & \textbf{65.1} \\
            & \textbf{YYDS+CMKR} & \textbf{ours} & \textbf{96.6} & \textbf{98.9} & \textbf{99.6} & \textbf{96.1} 
                                               & \textbf{73.4} & \textbf{91.3} & \textbf{94.6} & \textbf{77.4} \\                
            \bottomrule[1pt]
        \end{tabular}
        }
    \end{center}
  \end{table*}

\section{Experiments}
\label{sec:experiment}

\subsection{Datasets and Evaluation Protocol}

\textbf{SYSU-MM01} \cite{wu2017rgb} is the first large-scale benchmark dataset for VI-ReID, 
which is collected by 4 visible and 2 infrared cameras in both indoor and outdoor environments.
The training set contains 395 identities, including 22,258 visible images and 11,909 infrared images.
The test set contains 96 identities, with 3,803 infrared images for query and 301/3010 (single-shot / multi-shot) randomly selected visible images as the gallery set.
Meanwhile, it contains two test modes, i.e., all-search and indoor-search modes.

\noindent \textbf{RegDB} \cite{nguyen2017person} is a harsh lighting dataset, which is collected by a dual-camera system.
There are 206 identities for training and 206 identities for testing.
Each person has 10 visible images and 10 infrared images.
It contains two test modes, i.e., visible-to-infrared and infrared-to-visible modes.

\noindent \textbf{LLCM} \cite{zhang2023diverse} is a recently proposed low-light VI-ReID dataset, which is collected by 9 cameras in low-light environments.
The training set contains 30,921 images of 713 identities, and the test set contains 13,909 images of 351 identities.
Both visible-to-infrared and infrared-to-visible modes are used for evaluation.

\noindent \textbf{Evaluation Protocol.} We evaluate our model on the 10 trials with different training/testing splits to achieve stable performance following previous works \cite{ye2020dynamic, zhang2023diverse}.
The cumulative matching characteristics (CMC) \cite{moon2001computational} and mean average precision (mAP) are adopted as evaluation metrics.

\subsection{Implementation Details}

\textbf{YYDS.}
We implement YYDS with PyTorch \cite{paszke2019pytorch} on 2 NVIDIA Tesla T4 GPUs.
While training, each training mini-batch consists of 4 identities, and each identity contains 4 visible images and 4 infrared images.
We utilize DEEN \cite{zhang2023diverse} with pretrained ResNet-50 \cite{he2016deep} on ImageNet \cite{russakovsky2015imagenet} as baseline.
All input images are resized to 3 $\times$ 384 $\times$ 144, and several data augmentation techniques are adopted,
i.e., random gray scale, random cropping, random horizontal flipping and random erasing \cite{zhong2020random}.
The overall framework is trained for 80 epochs by SGD optimizer with momentum parameter $p$=0.9 \cite{qian1999momentum}.
The learning rate gradually rises up to $\eta$=0.1 by the warm-up scheme for the first 10 epochs,
and decays by a factor of 10 at the 20th and 50th epochs.
The hyper-parameters in Eq.\ref{loss_all} $\lambda_t$, $\lambda_j$ and $\lambda_c$ are set to 1, 1 and 1.

\noindent \textbf{CMKR.}
While testing, the Jaccard distance $d_{Jacc}$ is combined with original feature distance to perform re-ranking with a weight $\lambda_{Jacc}$.
For fair comparison, the best hyper-parameters are selected for each neighbor strategy respectively by grid search.

\subsection{Comparison with State-of-the-art Methods}

We compare the proposed ``YYDS+CMKR'' with the state-of-the-art VI-ReID methods on one excellent lighting dataset SYSU-MM01 and two harsh lighting datasets RegDB and LLCM.
The results are shown in Tab.\ref{table_sota_1} and Tab.\ref{table_sota_2}, and our methods outperform the other methods on all benchmarks by a large margin.
On SYSU-MM01, YYDS achieves Rank1 85.54\% and mAP 81.64\% in the all-search mode, higher than STAR by 9.47\% and 8.91\%.
CMKR further improves the performance to Rank1 95.51\% and mAP 93.77\%.
On RegDB and LLCM, ``YYDS+CMKR'' achieves better scores than DEEN by 7.1\% and 18.5\% for Rank1, and 12.7\% and 14.5\% for mAP.
This proves the effectiveness of our proposed network and re-ranking algorithm.

\begin{table*}[t]
    \renewcommand{\arraystretch}{1.2}
    \begin{center}
        \caption{
            \textbf{Ablation study on different components of YYDS on SYSU-MM01.}
            \textbf{Baseline:} Only texture encoders are utilized.
            \textbf{Color:} Both texture and color encoders are adopted.
            \textbf{IoU:} Text-IoU regularization is used.
            \textbf{Joint:} All three encoders are used.
        }
        \label{table_ablation_1}
        \resizebox{.9\textwidth}{!}{
        \begin{tabular}{cl|c|c|c|c|c|c|c}
            \toprule[1pt]
            & \textbf{   Method   } & \textbf{Color} & \textbf{IoU} & \textbf{Joint} & \textbf{R-1} & \textbf{R-10} & \textbf{R-20} & \textbf{mAP} \\
            \hline
            & \ Baseline &      -     &      -     &      -     & 72.36          & 96.56         & 99.03         & 68.24          \\
            &            & \checkmark &            &            & 83.66 (+11.30) & 99.22 (+2.66) & 99.83 (+0.80) & 78.90 (+10.66) \\
            &            & \checkmark & \checkmark &            & 84.54 (+12.18) & 98.84 (+2.28) & 99.63 (+0.60) & 80.01 (+11.77) \\
            &            & \checkmark & \checkmark & \checkmark & 85.54 (+13.18) & 99.30 (+2.74) & 99.78 (+0.75) & 81.64 (+13.40) \\
            \bottomrule[1pt]
        \end{tabular}
        }
    \end{center}
    \begin{center}
        \caption{
            \textbf{Ablation study on different components of JRM on SYSU-MM01.}
        }
        \label{table_ablation_2}
        \resizebox{.85\textwidth}{!}{
        \begin{tabular}{cl|c|c|c|c}
            \toprule[1pt]
            & \textbf{Method} & \textbf{R-1} & \textbf{R-10} & \textbf{R-20} & \textbf{mAP} \\
            \hline
            & YYDS                 & 85.54         & 99.30         & 99.78         & 81.64         \\
            & w/o Texture-centered Relation  & 84.63 (-0.91) & 98.81 (-0.49) & 99.70 (-0.08) & 81.43 (-0.21) \\
            & w/o Color-centered Relation & 84.63 (-0.91) & 98.53 (-0.77) & 99.56 (-0.22) & 81.23 (-0.41) \\
            & w/o Joint Relation   & 84.93 (-0.61) & 98.99 (-0.31) & 99.76 (-0.02) & 80.50 (-1.14) \\
            \bottomrule[1pt]
        \end{tabular}
        }
    \end{center}
    \begin{center}
        \caption{
            \textbf{Ablation study on different strategies of CMKR.}
            For fair comparison, the best hyperparameters are selected for each strategy with grid-search.
        }
        \label{table_ablation_3}
        \resizebox{.95\textwidth}{!}{
        \begin{tabular}{cl|c|c|c|c|c|c|c|c|c|c}
            \toprule[1pt]
            & \multirow{2}*{\textbf{Method}} 
            & \multirow{2}*{\textbf{constrained}} & \multirow{2}*{\textbf{\ \ divided\ \ }} & \multirow{2}*{\textbf{ extended }} & \multirow{2}*{\textbf{MA-LQE}}
            & \multicolumn{2}{c|}{\textbf{SYSU}} & \multicolumn{2}{c|}{\textbf{RegDB}} & \multicolumn{2}{c}{\textbf{LLCM}} \\
            \cline{7-12}
            & ~ & ~ & ~ & ~ & ~ & \textbf{R-1} & \textbf{mAP} & \textbf{R-1} & \textbf{mAP} & \textbf{R-1} & \textbf{mAP} \\
            \hline
            & YYDS                            & - & - & - & - & 85.54 & 81.64 & 90.16 & 83.52 & 58.22 & 65.09 \\
            & +re-ranking \cite{zhong2017re} & - & - & - & - & 95.24 & 93.19 & 92.18 & 91.82 & 72.63 & 76.61 \\
            &                                 & \checkmark & & & & 93.37 & 90.07 & 95.98 & 94.84 & 70.93 & 74.75 \\
            &                                 & & \checkmark & & & 95.40 & 93.23 & 96.73 & 94.47 & 73.39 & 77.13 \\
            &                                 & & & \checkmark & & \textbf{95.40} & \textbf{93.72} & \textbf{96.51} & \textbf{96.79} & \textbf{73.01} & \textbf{77.25} \\
            &                                 & & & \checkmark & \checkmark & \textbf{95.51} & \textbf{93.77} 
                                                                            & \textbf{96.65} & \textbf{96.13} & \textbf{73.41} & \textbf{77.35} \\
            \bottomrule[1pt]
        \end{tabular}
        }
    \end{center}
  \end{table*}  

\subsection{Ablation Study}

\noindent \textbf{YYDS.}
We conduct ablation studies of each component of YYDS in Tab.\ref{table_ablation_1}.
The introduction of color encoders improves the Rank1 and mAP by 11.30\% and 10.66\% respectively,
which proves the complementarity of color information introduced by coarse descriptions.
The text-IoU regularization is designed to alleviate the optimization confusion problem of color encoders,
and achieves improvements of Rank1 0.88\% and mAP 1.11\%.
Finally, the introduction of JRM further brings 1.00\% gains in Rank1 and 1.63\% gains in mAP.
We further study the contributions of three relation modules $E^*_{j,t}$, $E^*_{j,c}$ and $E^*_{j,tc}$ in JRM in Tab.\ref{table_ablation_2}.
The results prove the effectiveness of the asymmetric design of JRM.

\noindent \textbf{CMKR.}
We take k-reciprocal re-ranking \cite{zhong2017re} as baseline, and compare different neighbor strategies in Tab.\ref{table_ablation_3}.
The constrained strategy reduces performance on SYSU-MM01 and LLCM datasets, which is caused by losing intra-modal contextual information.
Instead, both divided and extended strategies achieve consistent improvements on all three datasets, and the latter is taken as the default setting of our CMKR.
Furthermore, MA-LQE solves the neighbor modality bias problem in neighbor features, and improves the Rank1 scores by 0.11\%, 0.14\% and 0.40\% on three datasets.

\begin{figure}[t]
    \centering
    \subfloat[Baseline]{\includegraphics[width=.45\textwidth]{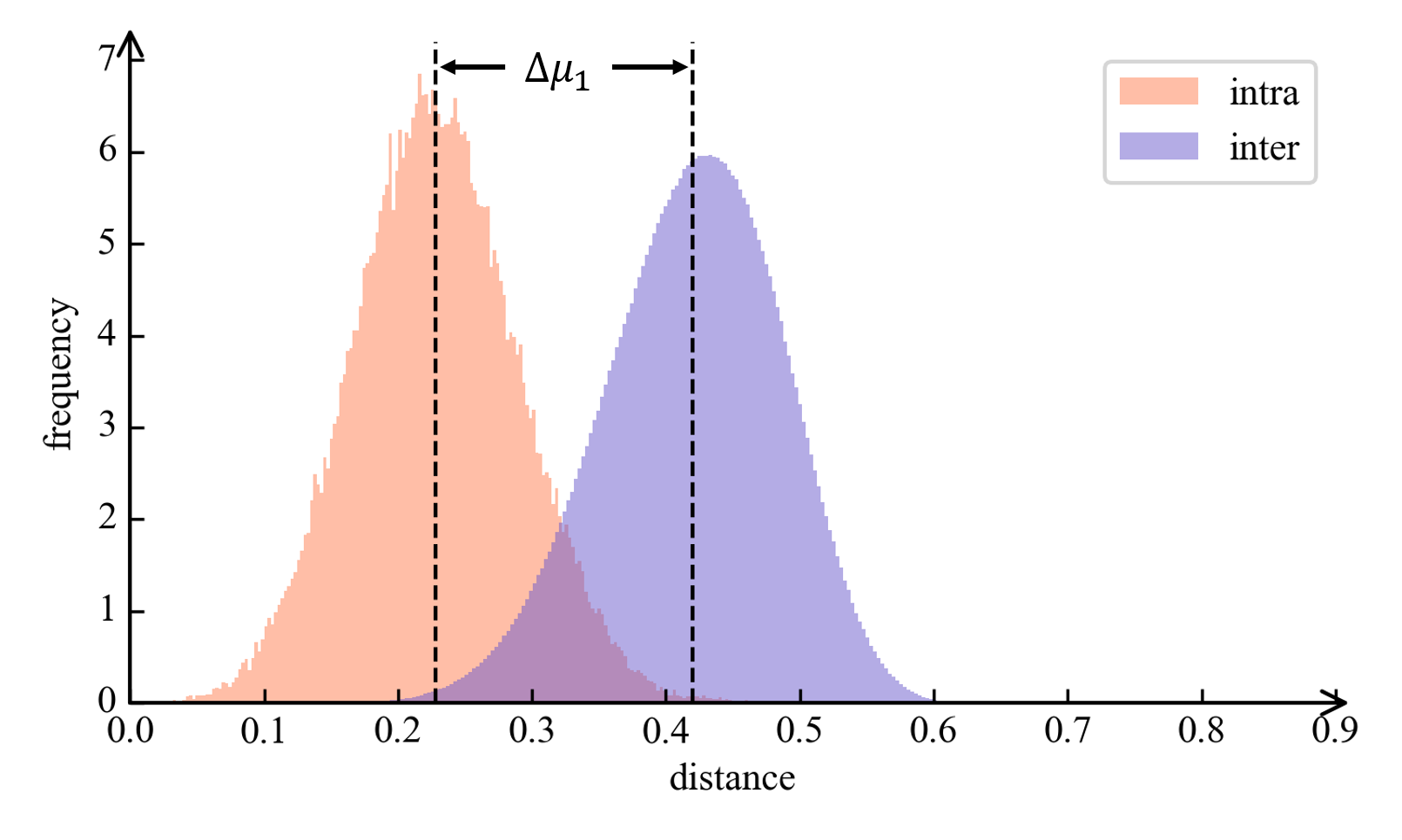}}
    \subfloat[YYDS]{\includegraphics[width=.45\textwidth]{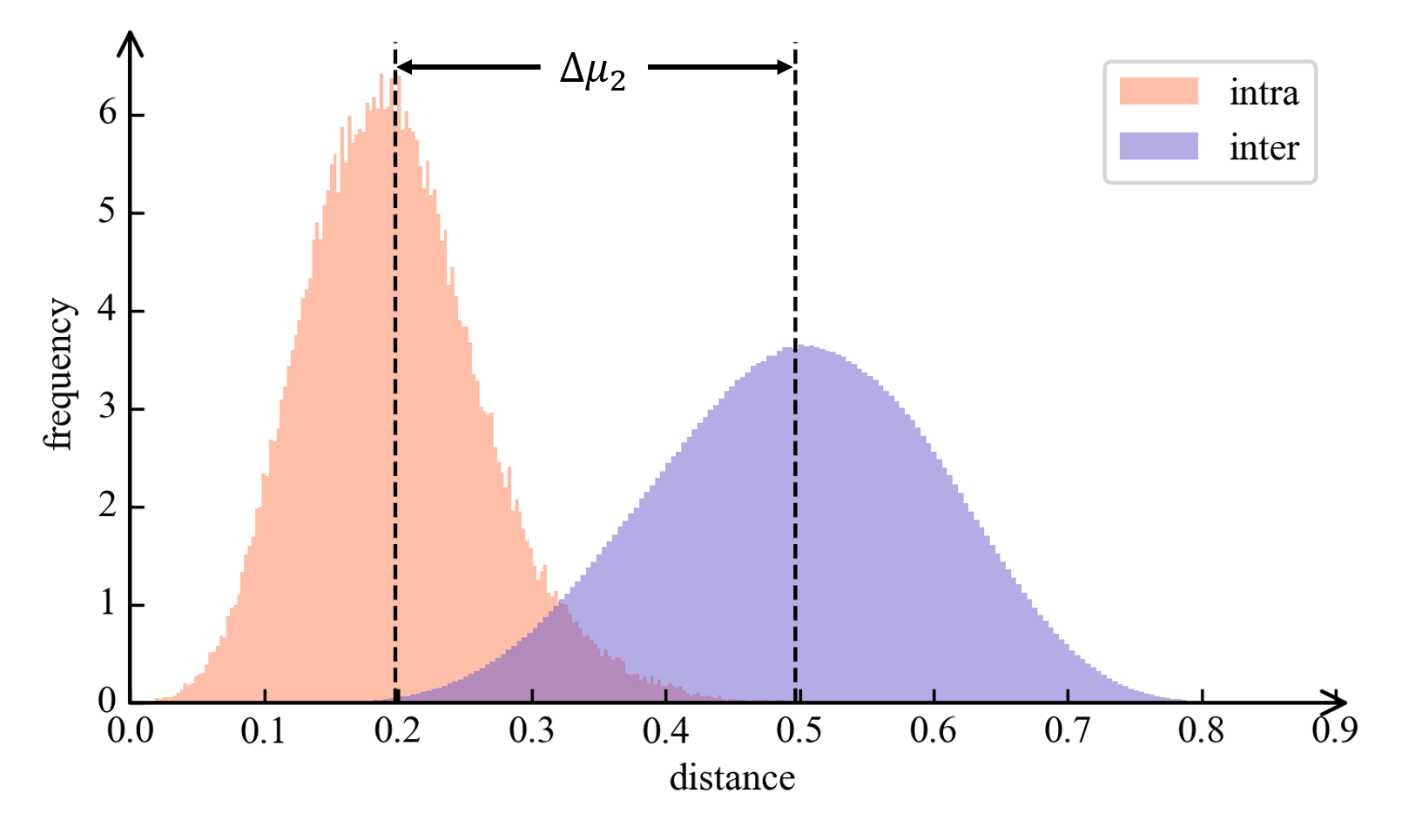}}
    \caption{
        \textbf{The intra-identity and inter-identity distribution of feature distances of baseline and YYDS.}
        (a) Baseline: $\Delta \mu_1 = 0.19$
        (b) YYDS: $\Delta \mu_2 = 0.30$.
    }
    \label{fig_distribution}
\end{figure}

\noindent \textbf{Visualization}
Fig.\ref{fig_distribution} visualize the feature distance distributions of baseline and YYDS on the SYSU-MM01 test set.
We can observe that our method can widen the gap $\Delta \mu$ between the mean distances of intra-identity and inter-identity.

\section{Conclusion}
In this paper, we present a new task setting, named Refer-VI-ReID, to make up for the missing color information in infrared images with coarse descriptions.
The main challenge lies in the misalignment between query modality and gallery modality.
To solve this problem, we propose a Y-Y-shape decomposition structure (YYDS),
which first decomposes the discriminative information into texture and color embeddings, and then aggregates them to the final features.
Specially, an elaborately designed joint relation module is utilized to perform asymmetric relation and dynamic aggregation.
While training, a text-IoU regularization strategy is introduced to solve the optimization confusion problem.

Considering that commonly used k-reciprocal re-ranking faces the neighbor modality bias problem in cross-modal scenarios,
we thoroughly study three neighbor strategies and one query expansion strategy, and propose the cross-modal k-reciprocal re-ranking (CMKR) algorithm. 
Experimental results on SYSU-MM01, RegDB and LLCM prove the effectiveness of our proposed methods.

\clearpage  

%
%
\bibliographystyle{splncs04}
\bibliography{main}

\begin{thebibliography}{100}
\providecommand{\url}[1]{\texttt{#1}}
\providecommand{\urlprefix}{URL }
\providecommand{\doi}[1]{https://doi.org/#1}

\bibitem{bai2016sparse}
Bai, S., Bai, X.: Sparse contextual activation for efficient visual re-ranking.
  IEEE Transactions on Image Processing  \textbf{25}(3),  1056--1069 (2016)

\bibitem{basaran2020efficient}
Basaran, E., G{\"o}kmen, M., Kamasak, M.E.: An efficient framework for
  visible--infrared cross modality person re-identification. Signal Processing:
  Image Communication  \textbf{87},  115933 (2020)

\bibitem{bird2009natural}
Bird, S., Klein, E., Loper, E.: Natural language processing with Python:
  analyzing text with the natural language toolkit. " O'Reilly Media, Inc."
  (2009)

\bibitem{chai2023dual}
Chai, Z., Ling, Y., Luo, Z., Lin, D., Jiang, M., Li, S.: Dual-stream
  transformer with distribution alignment for visible-infrared person
  re-identification. IEEE Transactions on Circuits and Systems for Video
  Technology  (2023)

\bibitem{chen2022structure}
Chen, C., Ye, M., Qi, M., Wu, J., Jiang, J., Lin, C.W.: Structure-aware
  positional transformer for visible-infrared person re-identification. IEEE
  Transactions on Image Processing  \textbf{31},  2352--2364 (2022)

\bibitem{chen2021explainable}
Chen, X., Liu, X., Liu, W., Zhang, X.P., Zhang, Y., Mei, T.: Explainable person
  re-identification with attribute-guided metric distillation. In: Proceedings
  of the IEEE/CVF International Conference on Computer Vision. pp. 11813--11822
  (2021)

\bibitem{chen2021neural}
Chen, Y., Wan, L., Li, Z., Jing, Q., Sun, Z.: Neural feature search for
  rgb-infrared person re-identification. In: Proceedings of the IEEE/CVF
  Conference on Computer Vision and Pattern Recognition. pp. 587--597 (2021)

\bibitem{chen2023jaccard}
Chen, Y., Fan, Z., Chen, Z., Zhu, Y.: Ca-jaccard: Camera-aware jaccard distance
  for person re-identification. arXiv preprint arXiv:2311.10605  (2023)

\bibitem{choi2020hierarchical}
Choi, S., Lee, S., Kim, Y., Kim, T., Kim, C.H.C.: Hierarchical cross-modality
  disentanglement for visible-infrared person re-identification. In:
  Proceedings of the2020 IEEE/CVF Conference Computer Vision Pattern
  Recognition, Seattle, WA, USA. pp. 13--19 (2020)

\bibitem{choi2020hi}
Choi, S., Lee, S., Kim, Y., Kim, T., Kim, C.: Hi-cmd: Hierarchical
  cross-modality disentanglement for visible-infrared person re-identification.
  In: Proceedings of the IEEE/CVF conference on computer vision and pattern
  recognition. pp. 10257--10266 (2020)

\bibitem{chollet2017xception}
Chollet, F.: Xception: Deep learning with depthwise separable convolutions. In:
  Proceedings of the IEEE conference on computer vision and pattern
  recognition. pp. 1251--1258 (2017)

\bibitem{chum2007total}
Chum, O., Philbin, J., Sivic, J., Isard, M., Zisserman, A.: Total recall:
  Automatic query expansion with a generative feature model for object
  retrieval. In: 2007 IEEE 11th International Conference on Computer Vision.
  pp.~1--8. IEEE (2007)

\bibitem{chun2021probabilistic}
Chun, S., Oh, S.J., De~Rezende, R.S., Kalantidis, Y., Larlus, D.: Probabilistic
  embeddings for cross-modal retrieval. In: Proceedings of the IEEE/CVF
  Conference on Computer Vision and Pattern Recognition. pp. 8415--8424 (2021)

\bibitem{dai2018cross}
Dai, P., Ji, R., Wang, H., Wu, Q., Huang, Y.: Cross-modality person
  re-identification with generative adversarial training. In: IJCAI. vol.~1,
  p.~6 (2018)

\bibitem{du2023enhanced}
Du, G., Zhang, L.: Enhanced invariant feature joint learning via
  modality-invariant neighbor relations for cross-modality person
  re-identification. IEEE Transactions on Circuits and Systems for Video
  Technology  (2023)

\bibitem{du2023video}
Du, Y., Lei, C., Zhao, Z., Dong, Y., Su, F.: Video-based visible-infrared
  person re-identification with auxiliary samples. IEEE Transactions on
  Information Forensics and Security  \textbf{19},  1313--1325 (2023)

\bibitem{du2022pami}
Du, Y., Tong, Z., Wan, J., Zhang, B., Zhao, Y.: Pami-ad: An activity detector
  exploiting part-attention and motion information in surveillance videos. In:
  2022 IEEE International Conference on Multimedia and Expo Workshops (ICMEW).
  pp.~1--6. IEEE (2022)

\bibitem{du2022omg}
Du, Y., Zhang, B., Ruan, X., Su, F., Zhao, Z., Chen, H.: Omg: Observe multiple
  granularities for natural language-based vehicle retrieval. In: Proceedings
  of the IEEE/CVF Conference on Computer Vision and Pattern Recognition. pp.
  3124--3133 (2022)

\bibitem{du2023strongsort}
Du, Y., Zhao, Z., Song, Y., Zhao, Y., Su, F., Gong, T., Meng, H.: Strongsort:
  Make deepsort great again. IEEE Transactions on Multimedia  (2023)

\bibitem{fan2022modality}
Fan, X., Jiang, W., Luo, H., Mao, W.: Modality-transfer generative adversarial
  network and dual-level unified latent representation for visible thermal
  person re-identification. The Visual Computer pp. 1--16 (2022)

\bibitem{fang2023visible}
Fang, X., Yang, Y., Fu, Y.: Visible-infrared person re-identification via
  semantic alignment and affinity inference. In: Proceedings of the IEEE/CVF
  International Conference on Computer Vision. pp. 11270--11279 (2023)

\bibitem{feng2023shape}
Feng, J., Wu, A., Zheng, W.S.: Shape-erased feature learning for
  visible-infrared person re-identification. In: Proceedings of the IEEE/CVF
  Conference on Computer Vision and Pattern Recognition. pp. 22752--22761
  (2023)

\bibitem{fu2021cm}
Fu, C., Hu, Y., Wu, X., Shi, H., Mei, T., He, R.: Cm-nas: Cross-modality neural
  architecture search for visible-infrared person re-identification. In:
  Proceedings of the IEEE/CVF International Conference on Computer Vision. pp.
  11823--11832 (2021)

\bibitem{gao2021contextual}
Gao, C., Cai, G., Jiang, X., Zheng, F., Zhang, J., Gong, Y., Peng, P., Guo, X.,
  Sun, X.: Contextual non-local alignment over full-scale representation for
  text-based person search. arXiv preprint arXiv:2101.03036  (2021)

\bibitem{garcia2015person}
Garcia, J., Martinel, N., Micheloni, C., Gardel, A.: Person re-identification
  ranking optimisation by discriminant context information analysis. In:
  Proceedings of the IEEE International Conference on Computer Vision. pp.
  1305--1313 (2015)

\bibitem{ge2018fd}
Ge, Y., Li, Z., Zhao, H., Yin, G., Yi, S., Wang, X., et~al.: Fd-gan:
  Pose-guided feature distilling gan for robust person re-identification.
  Advances in neural information processing systems  \textbf{31} (2018)

\bibitem{goodfellow2014generative}
Goodfellow, I., Pouget-Abadie, J., Mirza, M., Xu, B., Warde-Farley, D., Ozair,
  S., Courville, A., Bengio, Y.: Generative adversarial nets. Advances in
  neural information processing systems  \textbf{27} (2014)

\bibitem{gordo2017end}
Gordo, A., Almazan, J., Revaud, J., Larlus, D.: End-to-end learning of deep
  visual representations for image retrieval. International Journal of Computer
  Vision  \textbf{124}(2),  237--254 (2017)

\bibitem{guo2018density}
Guo, R.P., Li, C.G., Li, Y., Lin, J.: Density-adaptive kernel based re-ranking
  for person re-identification. In: 2018 24th International Conference on
  Pattern Recognition (ICPR). pp. 982--987. IEEE (2018)

\bibitem{hao2021cross}
Hao, X., Zhao, S., Ye, M., Shen, J.: Cross-modality person re-identification
  via modality confusion and center aggregation. In: Proceedings of the
  IEEE/CVF International conference on computer vision. pp. 16403--16412 (2021)

\bibitem{hao2019dual}
Hao, Y., Wang, N., Gao, X., Li, J., Wang, X.: Dual-alignment feature embedding
  for cross-modality person re-identification. In: Proceedings of the 27th ACM
  International Conference on Multimedia. pp. 57--65 (2019)

\bibitem{hao2019hsme}
Hao, Y., Wang, N., Li, J., Gao, X.: Hsme: Hypersphere manifold embedding for
  visible thermal person re-identification. In: Proceedings of the AAAI
  conference on artificial intelligence. vol.~33, pp. 8385--8392 (2019)

\bibitem{he2016deep}
He, K., Zhang, X., Ren, S., Sun, J.: Deep residual learning for image
  recognition. In: Proceedings of the IEEE conference on computer vision and
  pattern recognition. pp. 770--778 (2016)

\bibitem{he2023fastreid}
He, L., Liao, X., Liu, W., Liu, X., Cheng, P., Mei, T.: Fastreid: A pytorch
  toolbox for general instance re-identification. In: Proceedings of the 31st
  ACM International Conference on Multimedia. pp. 9664--9667 (2023)

\bibitem{he2021transreid}
He, S., Luo, H., Wang, P., Wang, F., Li, H., Jiang, W.: Transreid:
  Transformer-based object re-identification. In: Proceedings of the IEEE/CVF
  international conference on computer vision. pp. 15013--15022 (2021)

\bibitem{howard2017mobilenets}
Howard, A.G., Zhu, M., Chen, B., Kalenichenko, D., Wang, W., Weyand, T.,
  Andreetto, M., Adam, H.: Mobilenets: Efficient convolutional neural networks
  for mobile vision applications. arXiv preprint arXiv:1704.04861  (2017)

\bibitem{hu2020cross}
Hu, X., Zhou, Y.: Cross-modality person reid with maximum intra-class triplet
  loss. In: Pattern Recognition and Computer Vision: Third Chinese Conference,
  PRCV 2020, Nanjing, China, October 16--18, 2020, Proceedings, Part II 3. pp.
  557--568. Springer (2020)

\bibitem{huang2023vop}
Huang, S., Gong, B., Pan, Y., Jiang, J., Lv, Y., Li, Y., Wang, D.: Vop:
  Text-video co-operative prompt tuning for cross-modal retrieval. In:
  Proceedings of the IEEE/CVF Conference on Computer Vision and Pattern
  Recognition. pp. 6565--6574 (2023)

\bibitem{huang2021alleviating}
Huang, Y., Wu, Q., Xu, J., Zhong, Y., Zhang, P., Zhang, Z.: Alleviating
  modality bias training for infrared-visible person re-identification. IEEE
  Transactions on Multimedia  \textbf{24},  1570--1582 (2021)

\bibitem{huang2022modality}
Huang, Z., Liu, J., Li, L., Zheng, K., Zha, Z.J.: Modality-adaptive mixup and
  invariant decomposition for rgb-infrared person re-identification. In:
  Proceedings of the AAAI Conference on Artificial Intelligence. vol.~36, pp.
  1034--1042 (2022)

\bibitem{jambigi2021mmd}
Jambigi, C., Rawal, R., Chakraborty, A.: Mmd-reid: A simple but effective
  solution for visible-thermal person reid. arXiv preprint arXiv:2111.05059
  (2021)

\bibitem{jegou2007contextual}
Jegou, H., Harzallah, H., Schmid, C.: A contextual dissimilarity measure for
  accurate and efficient image search. In: 2007 IEEE Conference on computer
  vision and pattern recognition. pp.~1--8. IEEE (2007)

\bibitem{jegou2008accurate}
Jegou, H., Schmid, C., Harzallah, H., Verbeek, J.: Accurate image search using
  the contextual dissimilarity measure. IEEE Transactions on Pattern Analysis
  and Machine Intelligence  \textbf{32}(1),  2--11 (2008)

\bibitem{jia2020similarity}
Jia, M., Zhai, Y., Lu, S., Ma, S., Zhang, J.: A similarity inference metric for
  rgb-infrared cross-modality person re-identification. arXiv preprint
  arXiv:2007.01504  (2020)

\bibitem{jiang2023cross}
Jiang, D., Ye, M.: Cross-modal implicit relation reasoning and aligning for
  text-to-image person retrieval. In: Proceedings of the IEEE/CVF Conference on
  Computer Vision and Pattern Recognition. pp. 2787--2797 (2023)

\bibitem{jiang2022cross}
Jiang, K., Zhang, T., Liu, X., Qian, B., Zhang, Y., Wu, F.: Cross-modality
  transformer for visible-infrared person re-identification. In: European
  Conference on Computer Vision. pp. 480--496. Springer (2022)

\bibitem{kalayeh2018human}
Kalayeh, M.M., Basaran, E., G{\"o}kmen, M., Kamasak, M.E., Shah, M.: Human
  semantic parsing for person re-identification. In: Proceedings of the IEEE
  conference on computer vision and pattern recognition. pp. 1062--1071 (2018)

\bibitem{kim2023partmix}
Kim, M., Kim, S., Park, J., Park, S., Sohn, K.: Partmix: Regularization
  strategy to learn part discovery for visible-infrared person
  re-identification. In: Proceedings of the IEEE/CVF Conference on Computer
  Vision and Pattern Recognition. pp. 18621--18632 (2023)

\bibitem{kingma2013auto}
Kingma, D.P., Welling, M.: Auto-encoding variational bayes. arXiv preprint
  arXiv:1312.6114  (2013)

\bibitem{kong2021dynamic}
Kong, J., He, Q., Jiang, M., Liu, T.: Dynamic center aggregation loss with
  mixed modality for visible-infrared person re-identification. IEEE Signal
  Processing Letters  \textbf{28},  2003--2007 (2021)

\bibitem{kramer1991nonlinear}
Kramer, M.A.: Nonlinear principal component analysis using autoassociative
  neural networks. AIChE journal  \textbf{37}(2),  233--243 (1991)

\bibitem{leng2013bidirectional}
Leng, Q., Hu, R., Liang, C., Wang, Y., Chen, J.: Bidirectional ranking for
  person re-identification. In: 2013 IEEE International Conference on
  Multimedia and Expo (ICME). pp.~1--6. IEEE (2013)

\bibitem{leng2015person}
Leng, Q., Hu, R., Liang, C., Wang, Y., Chen, J.: Person re-identification with
  content and context re-ranking. Multimedia Tools and Applications
  \textbf{74},  6989--7014 (2015)

\bibitem{li2020infrared}
Li, D., Wei, X., Hong, X., Gong, Y.: Infrared-visible cross-modal person
  re-identification with an x modality. In: Proceedings of the AAAI conference
  on artificial intelligence. vol.~34, pp. 4610--4617 (2020)

\bibitem{li2023intermediary}
Li, H., Liu, M., Hu, Z., Nie, F., Yu, Z.: Intermediary-guided bidirectional
  spatial-temporal aggregation network for video-based visible-infrared person
  re-identification. IEEE Transactions on Circuits and Systems for Video
  Technology  (2023)

\bibitem{li2017person}
Li, S., Xiao, T., Li, H., Zhou, B., Yue, D., Wang, X.: Person search with
  natural language description. In: Proceedings of the IEEE conference on
  computer vision and pattern recognition. pp. 1970--1979 (2017)

\bibitem{li2023clip}
Li, S., Sun, L., Li, Q.: Clip-reid: exploiting vision-language model for image
  re-identification without concrete text labels. In: Proceedings of the AAAI
  Conference on Artificial Intelligence. vol.~37, pp. 1405--1413 (2023)

\bibitem{li2014deepreid}
Li, W., Zhao, R., Xiao, T., Wang, X.: Deepreid: Deep filter pairing neural
  network for person re-identification. In: Proceedings of the IEEE conference
  on computer vision and pattern recognition. pp. 152--159 (2014)

\bibitem{li2020gait}
Li, X., Makihara, Y., Xu, C., Yagi, Y., Ren, M.: Gait recognition via
  semi-supervised disentangled representation learning to identity and
  covariate features. In: Proceedings of the IEEE/CVF Conference on Computer
  Vision and Pattern Recognition. pp. 13309--13319 (2020)

\bibitem{liang2021homogeneous}
Liang, W., Wang, G., Lai, J., Xie, X.: Homogeneous-to-heterogeneous:
  Unsupervised learning for rgb-infrared person re-identification. IEEE
  Transactions on Image Processing  \textbf{30},  6392--6407 (2021)

\bibitem{lin2022learning}
Lin, X., Li, J., Ma, Z., Li, H., Li, S., Xu, K., Lu, G., Zhang, D.: Learning
  modal-invariant and temporal-memory for video-based visible-infrared person
  re-identification. In: Proceedings of the IEEE/CVF Conference on Computer
  Vision and Pattern Recognition. pp. 20973--20982 (2022)

\bibitem{ling2020class}
Ling, Y., Zhong, Z., Luo, Z., Rota, P., Li, S., Sebe, N.: Class-aware modality
  mix and center-guided metric learning for visible-thermal person
  re-identification. In: Proceedings of the 28th ACM international conference
  on multimedia. pp. 889--897 (2020)

\bibitem{liu2020unity}
Liu, C., Chang, X., Shen, Y.D.: Unity style transfer for person
  re-identification. In: Proceedings of the IEEE/CVF conference on computer
  vision and pattern recognition. pp. 6887--6896 (2020)

\bibitem{liu2020parameter}
Liu, H., Tan, X., Zhou, X.: Parameter sharing exploration and hetero-center
  triplet loss for visible-thermal person re-identification. IEEE Transactions
  on Multimedia  \textbf{23},  4414--4425 (2020)

\bibitem{liu2021sfanet}
Liu, H., Ma, S., Xia, D., Li, S.: Sfanet: A spectrum-aware feature augmentation
  network for visible-infrared person reidentification. IEEE Transactions on
  Neural Networks and Learning Systems  (2021)

\bibitem{liu2019roberta}
Liu, Y., Ott, M., Goyal, N., Du, J., Joshi, M., Chen, D., Levy, O., Lewis, M.,
  Zettlemoyer, L., Stoyanov, V.: Roberta: A robustly optimized bert pretraining
  approach. arXiv preprint arXiv:1907.11692  (2019)

\bibitem{lu2023tri}
Lu, Z., Lin, R., Hu, H.: Tri-level modality-information disentanglement for
  visible-infrared person re-identification. IEEE Transactions on Multimedia
  (2023)

\bibitem{luo2019spectral}
Luo, C., Chen, Y., Wang, N., Zhang, Z.: Spectral feature transformation for
  person re-identification. In: Proceedings of the IEEE/CVF international
  conference on computer vision. pp. 4976--4985 (2019)

\bibitem{luo2019strong}
Luo, H., Jiang, W., Gu, Y., Liu, F., Liao, X., Lai, S., Gu, J.: A strong
  baseline and batch normalization neck for deep person re-identification. IEEE
  Transactions on Multimedia  \textbf{22}(10),  2597--2609 (2019)

\bibitem{luo2022clip4clip}
Luo, H., Ji, L., Zhong, M., Chen, Y., Lei, W., Duan, N., Li, T.: Clip4clip: An
  empirical study of clip for end to end video clip retrieval and captioning.
  Neurocomputing  \textbf{508},  293--304 (2022)

\bibitem{moon2001computational}
Moon, H., Phillips, P.J.: Computational and performance aspects of pca-based
  face-recognition algorithms. Perception  \textbf{30}(3),  303--321 (2001)

\bibitem{nguyen2017person}
Nguyen, D.T., Hong, H.G., Kim, K.W., Park, K.R.: Person recognition system
  based on a combination of body images from visible light and thermal cameras.
  Sensors  \textbf{17}(3), ~605 (2017)

\bibitem{niu2023improving}
Niu, K., Huang, T., Huang, L., Wang, L., Zhang, Y.: Improving inconspicuous
  attributes modeling for person search by language. IEEE transactions on image
  processing  (2023)

\bibitem{ouyang2021contextual}
Ouyang, J., Wu, H., Wang, M., Zhou, W., Li, H.: Contextual similarity
  aggregation with self-attention for visual re-ranking. Advances in Neural
  Information Processing Systems  \textbf{34},  3135--3148 (2021)

\bibitem{park2021learning}
Park, H., Lee, S., Lee, J., Ham, B.: Learning by aligning: Visible-infrared
  person re-identification using cross-modal correspondences. In: Proceedings
  of the IEEE/CVF international conference on computer vision. pp. 12046--12055
  (2021)

\bibitem{paszke2019pytorch}
Paszke, A., Gross, S., Massa, F., Lerer, A., Bradbury, J., Chanan, G., Killeen,
  T., Lin, Z., Gimelshein, N., Antiga, L., et~al.: Pytorch: An imperative
  style, high-performance deep learning library. Advances in neural information
  processing systems  \textbf{32} (2019)

\bibitem{peng2019re}
Peng, C., Wang, N., Li, J., Gao, X.: Re-ranking high-dimensional deep local
  representation for nir-vis face recognition. IEEE Transactions on Image
  Processing  \textbf{28}(9),  4553--4565 (2019)

\bibitem{pu2020dual}
Pu, N., Chen, W., Liu, Y., Bakker, E.M., Lew, M.S.: Dual gaussian-based
  variational subspace disentanglement for visible-infrared person
  re-identification. In: Proceedings of the 28th ACM International Conference
  on Multimedia. pp. 2149--2158 (2020)

\bibitem{qi2023generative}
Qi, J., Liang, T., Liu, W., Li, Y., Jin, Y.: A generative-based image fusion
  strategy for visible-infrared person re-identification. IEEE Transactions on
  Circuits and Systems for Video Technology  (2023)

\bibitem{qian1999momentum}
Qian, N.: On the momentum term in gradient descent learning algorithms. Neural
  networks  \textbf{12}(1),  145--151 (1999)

\bibitem{qin2011hello}
Qin, D., Gammeter, S., Bossard, L., Quack, T., Van~Gool, L.: Hello neighbor:
  Accurate object retrieval with k-reciprocal nearest neighbors. In: CVPR 2011.
  pp. 777--784. IEEE (2011)

\bibitem{radenovic2018fine}
Radenovi{\'c}, F., Tolias, G., Chum, O.: Fine-tuning cnn image retrieval with
  no human annotation. IEEE transactions on pattern analysis and machine
  intelligence  \textbf{41}(7),  1655--1668 (2018)

\bibitem{russakovsky2015imagenet}
Russakovsky, O., Deng, J., Su, H., Krause, J., Satheesh, S., Ma, S., Huang, Z.,
  Karpathy, A., Khosla, A., Bernstein, M., et~al.: Imagenet large scale visual
  recognition challenge. International journal of computer vision
  \textbf{115},  211--252 (2015)

\bibitem{sarfraz2018pose}
Sarfraz, M.S., Schumann, A., Eberle, A., Stiefelhagen, R.: A pose-sensitive
  embedding for person re-identification with expanded cross neighborhood
  re-ranking. In: Proceedings of the IEEE conference on computer vision and
  pattern recognition. pp. 420--429 (2018)

\bibitem{shao2022learning}
Shao, Z., Zhang, X., Fang, M., Lin, Z., Wang, J., Ding, C.: Learning
  granularity-unified representations for text-to-image person
  re-identification. In: Proceedings of the 30th ACM International Conference
  on Multimedia. pp. 5566--5574 (2022)

\bibitem{shen2021re}
Shen, X., Xiao, Y., Hu, S.X., Sbai, O., Aubry, M.: Re-ranking for image
  retrieval and transductive few-shot classification. Advances in Neural
  Information Processing Systems  \textbf{34},  25932--25943 (2021)

\bibitem{sifre2014rigid}
Sifre, L., Mallat, S.: Rigid-motion scattering for texture classification.
  arXiv preprint arXiv:1403.1687  (2014)

\bibitem{sun2018beyond}
Sun, Y., Zheng, L., Yang, Y., Tian, Q., Wang, S.: Beyond part models: Person
  retrieval with refined part pooling (and a strong convolutional baseline).
  In: Proceedings of the European conference on computer vision (ECCV). pp.
  480--496 (2018)

\bibitem{tan2021incomplete}
Tan, H., Liu, X., Bian, Y., Wang, H., Yin, B.: Incomplete descriptor mining
  with elastic loss for person re-identification. IEEE Transactions on Circuits
  and Systems for Video Technology  \textbf{32}(1),  160--171 (2021)

\bibitem{wang2020cross}
Wang, G.A., Zhang, T., Yang, Y., Cheng, J., Chang, J., Liang, X., Hou, Z.G.:
  Cross-modality paired-images generation for rgb-infrared person
  re-identification. In: Proceedings of the AAAI conference on artificial
  intelligence. vol.~34, pp. 12144--12151 (2020)

\bibitem{wang2019rgb}
Wang, G., Zhang, T., Cheng, J., Liu, S., Yang, Y., Hou, Z.: Rgb-infrared
  cross-modality person re-identification via joint pixel and feature
  alignment. In: Proceedings of the IEEE/CVF International Conference on
  Computer Vision. pp. 3623--3632 (2019)

\bibitem{wang2022nformer}
Wang, H., Shen, J., Liu, Y., Gao, Y., Gavves, E.: Nformer: Robust person
  re-identification with neighbor transformer. In: Proceedings of the IEEE/CVF
  Conference on Computer Vision and Pattern Recognition. pp. 7297--7307 (2022)

\bibitem{wang2022optimal}
Wang, J., Zhang, Z., Chen, M., Zhang, Y., Wang, C., Sheng, B., Qu, Y., Xie, Y.:
  Optimal transport for label-efficient visible-infrared person
  re-identification. In: Computer Vision--ECCV 2022: 17th European Conference,
  Tel Aviv, Israel, October 23--27, 2022, Proceedings, Part XXIV. pp. 93--109.
  Springer (2022)

\bibitem{wang2018resource}
Wang, Y., Wang, L., You, Y., Zou, X., Chen, V., Li, S., Huang, G., Hariharan,
  B., Weinberger, K.Q.: Resource aware person re-identification across multiple
  resolutions. In: Proceedings of the IEEE conference on computer vision and
  pattern recognition. pp. 8042--8051 (2018)

\bibitem{wang2020vitaa}
Wang, Z., Fang, Z., Wang, J., Yang, Y.: Vitaa: Visual-textual attributes
  alignment in person search by natural language. In: Computer Vision--ECCV
  2020: 16th European Conference, Glasgow, UK, August 23--28, 2020,
  Proceedings, Part XII 16. pp. 402--420. Springer (2020)

\bibitem{wang2023multilateral}
Wang, Z., Gao, Z., Guo, K., Yang, Y., Wang, X., Shen, H.T.: Multilateral
  semantic relations modeling for image text retrieval. In: Proceedings of the
  IEEE/CVF Conference on Computer Vision and Pattern Recognition. pp.
  2830--2839 (2023)

\bibitem{wang2019learning}
Wang, Z., Wang, Z., Zheng, Y., Chuang, Y.Y., Satoh, S.: Learning to reduce
  dual-level discrepancy for infrared-visible person re-identification. In:
  Proceedings of the IEEE/CVF conference on computer vision and pattern
  recognition. pp. 618--626 (2019)

\bibitem{wei2020co}
Wei, X., Li, D., Hong, X., Ke, W., Gong, Y.: Co-attentive lifting for
  infrared-visible person re-identification. In: Proceedings of the 28th ACM
  international conference on multimedia. pp. 1028--1037 (2020)

\bibitem{wei2021syncretic}
Wei, Z., Yang, X., Wang, N., Gao, X.: Syncretic modality collaborative learning
  for visible infrared person re-identification. In: Proceedings of the
  IEEE/CVF International Conference on Computer Vision. pp. 225--234 (2021)

\bibitem{wei2023dual}
Wei, Z., Yang, X., Wang, N., Gao, X.: Dual-adversarial representation
  disentanglement for visible infrared person re-identification. IEEE
  Transactions on Information Forensics and Security  (2023)

\bibitem{wu2017rgb}
Wu, A., Zheng, W.S., Yu, H.X., Gong, S., Lai, J.: Rgb-infrared cross-modality
  person re-identification. In: Proceedings of the IEEE international
  conference on computer vision. pp. 5380--5389 (2017)

\bibitem{wu2023style}
Wu, J., Liu, H., Shi, W., Liu, M., Li, W.: Style-agnostic representation
  learning for visible-infrared person re-identification. IEEE Transactions on
  Multimedia  (2023)

\bibitem{wu2023learning}
Wu, J., Liu, H., Su, Y., Shi, W., Tang, H.: Learning concordant attention via
  target-aware alignment for visible-infrared person re-identification. In:
  Proceedings of the IEEE/CVF International Conference on Computer Vision. pp.
  11122--11131 (2023)

\bibitem{wu2021discover}
Wu, Q., Dai, P., Chen, J., Lin, C.W., Wu, Y., Huang, F., Zhong, B., Ji, R.:
  Discover cross-modality nuances for visible-infrared person
  re-identification. In: Proceedings of the IEEE/CVF Conference on Computer
  Vision and Pattern Recognition. pp. 4330--4339 (2021)

\bibitem{yan2023clip}
Yan, S., Dong, N., Zhang, L., Tang, J.: Clip-driven fine-grained text-image
  person re-identification. IEEE Transactions on Image Processing  (2023)

\bibitem{yang2020mining}
Yang, F., Wang, Z., Xiao, J., Satoh, S.: Mining on heterogeneous manifolds for
  zero-shot cross-modal image retrieval. In: Proceedings of the AAAI Conference
  on Artificial Intelligence. vol.~34, pp. 12589--12596 (2020)

\bibitem{yang2022learning}
Yang, M., Huang, Z., Hu, P., Li, T., Lv, J., Peng, X.: Learning with twin noisy
  labels for visible-infrared person re-identification. In: Proceedings of the
  IEEE/CVF conference on computer vision and pattern recognition. pp.
  14308--14317 (2022)

\bibitem{ye2020bi}
Ye, H., Liu, H., Meng, F., Li, X.: Bi-directional exponential angular triplet
  loss for rgb-infrared person re-identification. IEEE Transactions on Image
  Processing  \textbf{30},  1583--1595 (2020)

\bibitem{ye2015coupled}
Ye, M., Chen, J., Leng, Q., Liang, C., Wang, Z., Sun, K.: Coupled-view based
  ranking optimization for person re-identification. In: MultiMedia Modeling:
  21st International Conference, MMM 2015, Sydney, NSW, Australia, January 5-7,
  2015, Proceedings, Part I 21. pp. 105--117. Springer (2015)

\bibitem{ye2019modality}
Ye, M., Lan, X., Leng, Q.: Modality-aware collaborative learning for visible
  thermal person re-identification. In: Proceedings of the 27th ACM
  International Conference on Multimedia. pp. 347--355 (2019)

\bibitem{ye2020cross}
Ye, M., Lan, X., Leng, Q., Shen, J.: Cross-modality person re-identification
  via modality-aware collaborative ensemble learning. IEEE Transactions on
  Image Processing  \textbf{29},  9387--9399 (2020)

\bibitem{ye2018hierarchical}
Ye, M., Lan, X., Li, J., Yuen, P.: Hierarchical discriminative learning for
  visible thermal person re-identification. In: Proceedings of the AAAI
  Conference on Artificial Intelligence. vol.~32 (2018)

\bibitem{ye2019bi}
Ye, M., Lan, X., Wang, Z., Yuen, P.C.: Bi-directional center-constrained
  top-ranking for visible thermal person re-identification. IEEE Transactions
  on Information Forensics and Security  \textbf{15},  407--419 (2019)

\bibitem{ye2016person}
Ye, M., Liang, C., Yu, Y., Wang, Z., Leng, Q., Xiao, C., Chen, J., Hu, R.:
  Person reidentification via ranking aggregation of similarity pulling and
  dissimilarity pushing. IEEE Transactions on Multimedia  \textbf{18}(12),
  2553--2566 (2016)

\bibitem{ye2021channel}
Ye, M., Ruan, W., Du, B., Shou, M.Z.: Channel augmented joint learning for
  visible-infrared recognition. In: Proceedings of the IEEE/CVF International
  Conference on Computer Vision. pp. 13567--13576 (2021)

\bibitem{ye2020dynamic}
Ye, M., Shen, J., J.~Crandall, D., Shao, L., Luo, J.: Dynamic dual-attentive
  aggregation learning for visible-infrared person re-identification. In:
  Computer Vision--ECCV 2020: 16th European Conference, Glasgow, UK, August
  23--28, 2020, Proceedings, Part XVII 16. pp. 229--247. Springer (2020)

\bibitem{ye2021deep}
Ye, M., Shen, J., Lin, G., Xiang, T., Shao, L., Hoi, S.C.: Deep learning for
  person re-identification: A survey and outlook. IEEE transactions on pattern
  analysis and machine intelligence  \textbf{44}(6),  2872--2893 (2021)

\bibitem{ye2020visible}
Ye, M., Shen, J., Shao, L.: Visible-infrared person re-identification via
  homogeneous augmented tri-modal learning. IEEE Transactions on Information
  Forensics and Security  \textbf{16},  728--739 (2020)

\bibitem{ye2018visible}
Ye, M., Wang, Z., Lan, X., Yuen, P.C.: Visible thermal person re-identification
  via dual-constrained top-ranking. In: IJCAI. vol.~1, p.~2 (2018)

\bibitem{yu2023modality}
Yu, H., Cheng, X., Peng, W., Liu, W., Zhao, G.: Modality unifying network for
  visible-infrared person re-identification. In: Proceedings of the IEEE/CVF
  International Conference on Computer Vision. pp. 11185--11195 (2023)

\bibitem{yu2017divide}
Yu, R., Zhou, Z., Bai, S., Bai, X.: Divide and fuse: A re-ranking approach for
  person re-identification. arXiv preprint arXiv:1708.04169  (2017)

\bibitem{zhang2023protohpe}
Zhang, G., Zhang, Y., Tan, Z.: Protohpe: Prototype-guided high-frequency patch
  enhancement for visible-infrared person re-identification. In: Proceedings of
  the 31st ACM International Conference on Multimedia. pp. 944--954 (2023)

\bibitem{zhang2022fmcnet}
Zhang, Q., Lai, C., Liu, J., Huang, N., Han, J.: Fmcnet: Feature-level modality
  compensation for visible-infrared person re-identification. In: Proceedings
  of the IEEE/CVF Conference on Computer Vision and Pattern Recognition. pp.
  7349--7358 (2022)

\bibitem{zhang2020deep}
Zhang, S., Chen, C., Song, W., Gan, Z.: Deep feature learning with attributes
  for cross-modality person re-identification. Journal of Electronic Imaging
  \textbf{29}(3),  033017--033017 (2020)

\bibitem{zhang2018deep}
Zhang, Y., Lu, H.: Deep cross-modal projection learning for image-text
  matching. In: Proceedings of the European conference on computer vision
  (ECCV). pp. 686--701 (2018)

\bibitem{zhang2022modality}
Zhang, Y., Zhao, S., Kang, Y., Shen, J.: Modality synergy complement learning
  with cascaded aggregation for visible-infrared person re-identification. In:
  European Conference on Computer Vision. pp. 462--479. Springer (2022)

\bibitem{zhang2023diverse}
Zhang, Y., Wang, H.: Diverse embedding expansion network and low-light
  cross-modality benchmark for visible-infrared person re-identification. In:
  Proceedings of the IEEE/CVF Conference on Computer Vision and Pattern
  Recognition. pp. 2153--2162 (2023)

\bibitem{zhang2021towards}
Zhang, Y., Yan, Y., Lu, Y., Wang, H.: Towards a unified middle modality
  learning for visible-infrared person re-identification. In: Proceedings of
  the 29th ACM International Conference on Multimedia. pp. 788--796 (2021)

\bibitem{zhang2023graph}
Zhang, Y., Qian, Q., Wang, H., Liu, C., Chen, W., Wan, F.: Graph convolution
  based efficient re-ranking for visual retrieval. IEEE Transactions on
  Multimedia  (2023)

\bibitem{zhang2019gait}
Zhang, Z., Tran, L., Yin, X., Atoum, Y., Liu, X., Wan, J., Wang, N.: Gait
  recognition via disentangled representation learning. In: Proceedings of the
  IEEE/CVF conference on computer vision and pattern recognition. pp.
  4710--4719 (2019)

\bibitem{zhang2021rgb}
Zhang, Z., Jiang, S., Huang, C., Li, Y., Da~Xu, R.Y.: Rgb-ir cross-modality
  person reid based on teacher-student gan model. Pattern Recognition Letters
  \textbf{150},  155--161 (2021)

\bibitem{zheng2022progressive}
Zheng, A., Pan, P., Li, H., Li, C., Luo, B., Tan, C., Jia, R.: Progressive
  attribute embedding for accurate cross-modality person re-id. In: Proceedings
  of the 30th ACM International Conference on Multimedia. pp. 4309--4317 (2022)

\bibitem{zheng2017person}
Zheng, L., Zhang, H., Sun, S., Chandraker, M., Yang, Y., Tian, Q.: Person
  re-identification in the wild. In: Proceedings of the IEEE conference on
  computer vision and pattern recognition. pp. 1367--1376 (2017)

\bibitem{zheng2019joint}
Zheng, Z., Yang, X., Yu, Z., Zheng, L., Yang, Y., Kautz, J.: Joint
  discriminative and generative learning for person re-identification. In:
  proceedings of the IEEE/CVF conference on computer vision and pattern
  recognition. pp. 2138--2147 (2019)

\bibitem{zheng2020dual}
Zheng, Z., Zheng, L., Garrett, M., Yang, Y., Xu, M., Shen, Y.D.: Dual-path
  convolutional image-text embeddings with instance loss. ACM Transactions on
  Multimedia Computing, Communications, and Applications (TOMM)
  \textbf{16}(2),  1--23 (2020)

\bibitem{zhong2021grayscale}
Zhong, X., Lu, T., Huang, W., Ye, M., Jia, X., Lin, C.W.: Grayscale enhancement
  colorization network for visible-infrared person re-identification. IEEE
  Transactions on Circuits and Systems for Video Technology  \textbf{32}(3),
  1418--1430 (2021)

\bibitem{zhong2017re}
Zhong, Z., Zheng, L., Cao, D., Li, S.: Re-ranking person re-identification with
  k-reciprocal encoding. In: Proceedings of the IEEE conference on computer
  vision and pattern recognition. pp. 1318--1327 (2017)

\bibitem{zhong2020random}
Zhong, Z., Zheng, L., Kang, G., Li, S., Yang, Y.: Random erasing data
  augmentation. In: Proceedings of the AAAI conference on artificial
  intelligence. vol.~34, pp. 13001--13008 (2020)

\bibitem{zhong2018camera}
Zhong, Z., Zheng, L., Zheng, Z., Li, S., Yang, Y.: Camera style adaptation for
  person re-identification. In: Proceedings of the IEEE conference on computer
  vision and pattern recognition. pp. 5157--5166 (2018)

\bibitem{zhou2021learning}
Zhou, K., Yang, Y., Cavallaro, A., Xiang, T.: Learning generalisable omni-scale
  representations for person re-identification. IEEE transactions on pattern
  analysis and machine intelligence  \textbf{44}(9),  5056--5069 (2021)

\bibitem{zhu2020hetero}
Zhu, Y., Yang, Z., Wang, L., Zhao, S., Hu, X., Tao, D.: Hetero-center loss for
  cross-modality person re-identification. Neurocomputing  \textbf{386},
  97--109 (2020)

\end{thebibliography}
\end{document}